\documentclass[journal,twoside]{IEEEtran}
\usepackage{cite}
\usepackage{graphicx, times,psfig,amsmath}
\usepackage{graphicx}

\usepackage[ruled, lined, linesnumbered, commentsnumbered, longend]{algorithm2e}

\usepackage{float,epsfig}
\usepackage{afterpage}
\usepackage{multirow}
\usepackage{amsthm}
\usepackage{bm}
\usepackage{amssymb}
\usepackage{booktabs}
\usepackage{colortbl}
\usepackage{hyperref}
\usepackage{url}
\usepackage{ifpdf}
\usepackage{threeparttable}
\usepackage{color,xcolor}
\usepackage{subcaption}
\usepackage{verbatim}

\newcommand{\miqing}[1]{\textcolor{red}{ Miqing: #1}}

\begin{document}

\title{A Weight Adaptation Trigger Mechanism in Decomposition-based Evolutionary Multi-Objective Optimisation}

	


	

\author{Xiaofeng Han$^\star$, Xiaochen Chu$^\star$, Tao Chao, Ming Yang, and Miqing Li$^\dagger$ 
\thanks{$^\star$ These authors contributed equally to this work.}
\thanks{X. Han, T. Chao and M. Yang are with the Control and Simulation Center in the Department of Control Science and Engineering, Harbin Institute of Technology (HIT), Harbin, China. (e-mail: hanx63520@gmail.com, chaotao2000@163.com, myang@hit.edu.cn).}
\thanks{X. Chu is with the Centre for Control Theory and Guidance Technology in the Department of Control Science and Engineering, Harbin Institute of Technology (HIT), Harbin, China. (e-mail: chuxiaochen95@163.com).}
\thanks{$^\dagger$ Corresponding author: Miqing Li.}
\thanks{M. Li is with the School of Computer Science, University of Birmingham, Birmingham B15 2TT, U.~K. (e-mail: m.li.8@bham.ac.uk).}
}

\markboth{IEEE Transactions on xxxx,~Vol.~xx, No.~xx, Year.~xx}{
	\MakeLowercase{\textit{et al.}}: A Weight Adaptation Trigger Mechanism in Decomposition-based Evolutionary Multi-Objective Optimisation}


\maketitle

\begin{abstract}
Decomposition-based multi-objective evolutionary algorithms (MOEAs) are widely used for solving multi-objective optimisation problems. 
However, their effectiveness depends on the consistency between the problem's Pareto front shape and the weight distribution. Decomposition-based MOEAs, with uniformly distributed weights (in a simplex), perform well on problems with a regular (simplex-like) Pareto front, but not on those with an irregular Pareto front. 
Previous studies have focused on adapting the weights to approximate the irregular Pareto front during the evolutionary process. 
However, these adaptations can actually harm the performance on the regular Pareto front via changing the weights during the search process that are eventually
the best fit for the Pareto front.
In this paper, we propose an algorithm called the weight adaptation trigger mechanism for decomposition-based MOEAs (ATM-MOEA/D) to tackle this issue. ATM-MOEA/D uses an archive to gradually approximate the shape of the Pareto front during the search. 
When the algorithm detects evolution stagnation (meaning the population no longer improves significantly), it compares the distribution of the population with that of the archive to distinguish between regular and irregular Pareto fronts.
Only when an irregular Pareto front is identified, the weights are adapted. 
Our experimental results show that the proposed algorithm not only performs generally better than seven state-of-the-art weight-adapting methods on irregular Pareto fronts but also is able to achieve the same results as fixed-weight
methods like MOEA/D on regular Pareto fronts.

\end{abstract}

\markboth{IEEE Transactions on xxxx,~Vol.~xx, No.~xx, Year.~xx}
{A Weight Adaptation Triggering Mechanism in Decomposition-based Evolutionary Multi-Objective Optimisation}



\section{Introduction}

\IEEEPARstart{D}{ecomposition-based} multi-objective evolutionary algorithms (MOEAs) have been demonstrated to be a good option for solving multi-objective optimisation problems (MOPs)~\cite{Zhang2007}. Decomposition-based MOEAs achieve this by using a set of weights (also called reference points in some studies)
to decompose an MOP into several scalar optimisation sub-problems, which are then optimised in a cooperative manner. Each weight is associated with a sub-problem and ideally corresponds to a solution in the population.
\IEEEpubidadjcol

Typically, decomposition-based MOEAs can lead to well-distributed solutions on problems with a regular Pareto front but may fail on those with an irregular Pareto front. 
The reason is as follows. 
In decomposition-based MOEAs, the weights determine the search directions of solutions in the objective space and guide them to move towards the Pareto fronts during the evolutionary process. 
Ideally, the Pareto optimal solutions that these weights are associated with are located at the intersections between the weights directions and the problem's Pareto fronts. Therefore, the distribution of the solutions heavily depends on the consistency between the problem's Pareto front shape and the distribution of the weights. 
Without knowing the problem's Pareto front shape, decomposition-based MOEAs typically consider uniformly distributed weights in a simplex (a triangle plane or a quarter sphere in the tri-objective case). Consequently, decomposition-based MOEAs with such predefined weights perform well on regular (simplex-like) Pareto fronts but poorly on irregular Pareto fronts.

To improve the performance on irregular Pareto fronts, an intuitive idea is to adapt the weights based on the shape of the Pareto front~\cite{Ishibuchi2016, Ma2020, Hua2021}.
Many interesting studies have been conducted along this line~\cite{Deb2014, Qi2014, Cheng2016, jiang2015a, Wu2018, Liu2019, Liu2020, Li2020, Liu2021}. 
They adapt the weights during the evolutionary process in order to approximate the shape of the Pareto front, and have demonstrated effectiveness in improving the diversity of the solutions on various irregular Pareto fronts.

Yet, these weight adaptation methods may harm the performance on MOPs with regular Pareto fronts.
If the distribution of the weights is already well-matched with the problem's Pareto front shape,
adapting them during the search can lead to the fluctuation of the search~\cite{Tian2017, Asafuddoula2017, Ma2020}. 
Moreover, once the weights are changed, 
it may be difficult to precisely bring them back to their original positions (since the non-dominated front
of the evolutionary population is unlike to be the same as 
the problem's Pareto front. 
As a result, existing weight adaptation methods perform worse in terms of both convergence and diversity than the original decomposition-based MOEAs for regular Pareto fronts. 

To address this issue, a possible way is to differentiate MOPs with regular Pareto fronts from ones with irregular Pareto fronts during the evolutionary process, 
so that one can adapt the weights only for irregular Pareto fronts.
In this paper, we make an attempt along this line and propose a weight adaptation trigger mechanism, called ATM-MOEA/D. 
In ATM-MOEA/D, 
a crucial issue is the timing of adapting the weights.
Here we consider the state that the evolution tends to stagnate (i.e., the population no longer improves significantly).
At that state, by comparing the population with a well-maintained archive, one may know if the weights need to be adapted or not.

The remainder of this paper is structured as follows: 
In Section II, we introduce related work and give the motivation of our work. 
Section III describes details of the proposed ATM-MOEA/D, 
and in Section IV, we experimentally verify the effectiveness of ATM-MOEA/D by comparing it with state-of-the-art algorithms. 
Finally, Section V concludes the paper.

%

\section{Related Work and Motivation}

Over the past two decades, MOEAs have been developed and applied to multi-objective optimisation problems. 
These MOEAs can be categorised into three classes~\cite{Zhou2011,LiB2015}: Pareto-based~\cite{Deb2002}, indicator-based~\cite{Zitzler2004}, and decomposition-based MOEAs~\cite{Zhang2007}. Among the three classes, decomposition-based MOEAs have become increasingly popular for solving MOPs due to their clear strengths. These strengths include providing high selection pressure~\cite{LiK2014}, having evenly distributed solutions on certain problems~\cite{LiK2015}, and being capable of handling many-objective optimisation problems~\cite{Asafuddoula2015b, LiK2015, Yuan2015b}. 
\begin{figure*}[tbp]
	\begin{center}
        \begin{tabular}{@{}c@{}c@{}c@{}c@{}}			
			\includegraphics[scale=0.35]{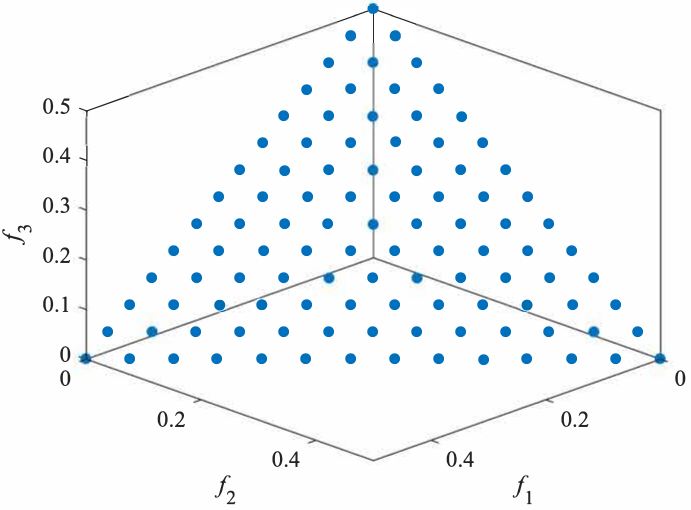}&
			\includegraphics[scale=0.35]{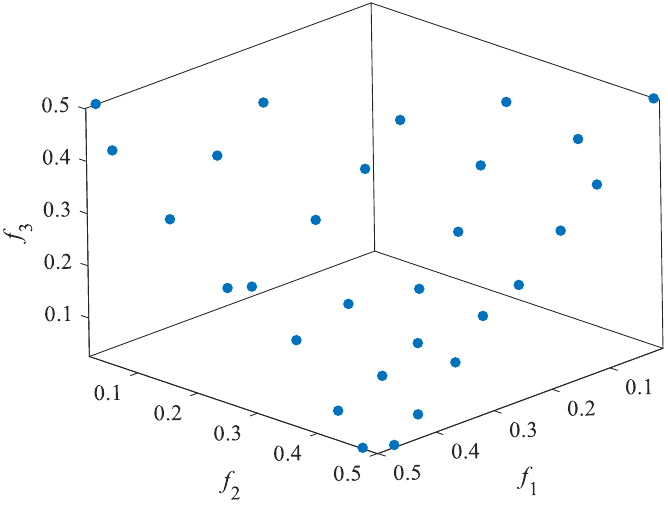}&
			\includegraphics[scale=0.35]{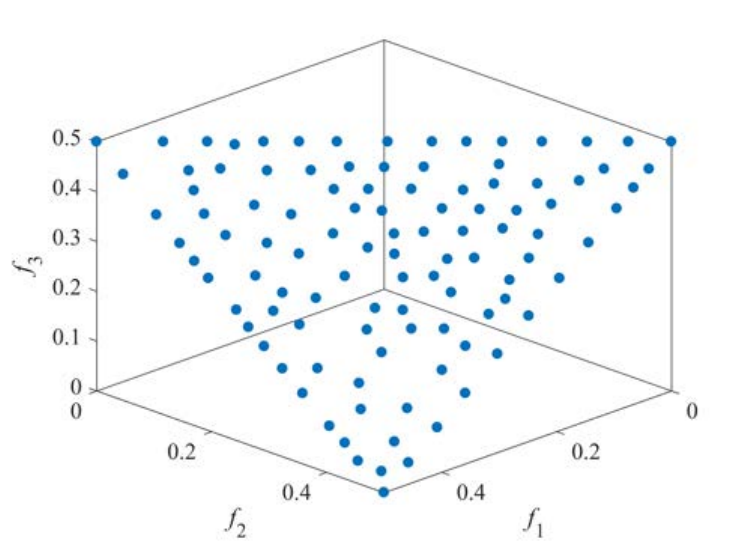}&
			\includegraphics[scale=0.35]{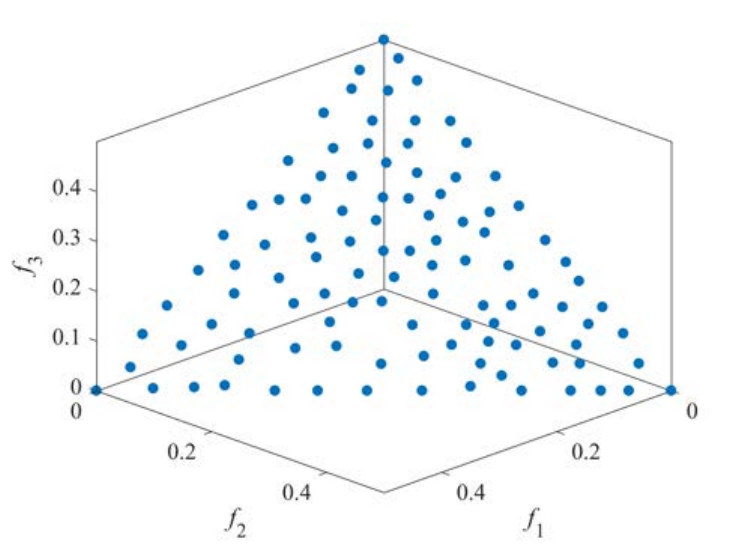}\\
			(a) RVEA on DTLZ1 & (b) RVEA on IDTLZ1 & (c) RVEAi-GNG on IDTLZ1 & (d) RVEAi-GNG on DTLZ1 \\
		\end{tabular}
	\end{center}
	\caption{Comparison of the performance of a weight-fixing algorithm, RVEA~\cite{Cheng2016}, and a weight-adaptive algorithm, RVEAi-GNG~\cite{Liu2020}, on two tri-objective optimisation problems: DTLZ1~\cite{Deb2005a} with a regular Pareto front and IDTLZ1~\cite{Deb2014} with an irregular Pareto front.}
	\label{fig:sec2}
\end{figure*}

Decomposition-based MOEAs solve an MOP by decomposing them into sub-problems using predefined weights and solving them collaboratively. The diversity of the solutions obtained is controlled by the weights, which can be initialised using different methods. They include simplex-lattice design~\cite{Das1998}, multi-layer simplex-lattice design~\cite{Deb2014, Jiang2017, Tan2013}, and uniform random sampling design~\cite{Ma2014}. These methods can generate a set of uniformly distributed weights.



However, uniformly distributed weights work well only for MOPs with regular Pareto fronts, where uniformly distributed solutions can be obtained. For MOPs with irregular Pareto fronts, the solutions obtained are likely to be distributed poorly~\cite{Ishibuchi2016, Ma2020, Hua2021,elarbi2019}. In such a case, several weights may correspond to one solution, wasting computational resources and more importantly, reducing the options (i.e., candidate solutions) for the decision maker in the decision-making process. 
Table~\ref{tab:Sec2} (upper part) lists several decomposition-based MOEAs with uniformly distributed
weights such as MOEA/D~\cite{Zhang2007}, MOEA/D-DE~\cite{Li2009a}, NSGA-III~\cite{Deb2014}, MOEA/D-M2M~\cite{Liu2013} and RVEA~\cite{Cheng2016}. These algorithms are effective when the problem's Pareto front is regular, but are ineffective when the problem's Pareto front is irregular~\cite{Deb2014, Cheng2016, Li2019, Liu2020, Qi2014, Li2020}.

\begin{table}[htbp]
  \centering
  \caption{Performance of representative decomposition-based MOEAs for regular and irregular Pareto fronts.}
    \begin{tabular}{p{2.6cm}m{1cm}<{\centering}m{1.6cm}<{\centering}m{1.7cm}<{\centering}}
    \hline
    \specialrule{0em}{1pt}{1pt}
    \multirow{2}{2.6cm}{Algorithm} & \multirow{2}{1cm}{\centering Type} & \multicolumn{2}{c}{Pareto fronts} \\
          &       & Regular & Irregular \\
    \specialrule{0em}{1pt}{1pt}
    \hline
    \specialrule{0em}{1pt}{1pt}
    MOEA/D~\cite{Zhang2007} & \multirow{5}{1cm}{\centering Fixed weights} & \multirow{5}{1.6cm}{\centering Excellent} & \multirow{5}{1.7cm}{\centering Poor} \\
    MOEA/D-DE~\cite{Li2009a} &       &       &  \\
    NSGA-III~\cite{Deb2014} &       &       &  \\
    MOEA/D-M2M~\cite{Liu2013} &       &       &  \\
    RVEA~\cite{Cheng2016}  &       &       &  \\
    \specialrule{0em}{1pt}{1pt}
    \hline
    \specialrule{0em}{1pt}{1pt}
    MOEA/D-AWA~\cite{Qi2014} & \multirow{10}{1cm}{\centering Adaptive weights} & \multirow{10}{1.6cm}{\centering Worse than weight-fixing algorithms} & \multirow{10}{1.7cm}{\centering Better than weight-fixing algorithms} \\
    iRVEA~\cite{Liu2019b} &       &       &  \\
    MOEA/D-LTD~\cite{Wu2018} &       &       &  \\
    DEA-GNG~\cite{Liu2019} &       &       &  \\
    RVEA-iGNG~\cite{Liu2020} &       &       &  \\
    AdaW~\cite{Li2020}  &       &       &  \\
    MOEA/D-AM2M~\cite{Liu2017} &       &       &  \\
    A-NSGA-III~\cite{Deb2014} &       &       &  \\
    RVEA*~\cite{Cheng2016} &       &       &  \\
    g-DBEA~\cite{Asafuddoula2017} &       &       &  \\
    \specialrule{0em}{1pt}{1pt}
    \hline
    \end{tabular}%
  \label{tab:Sec2}%
\end{table}%

For example, 
Figure~\ref{fig:sec2}(a) and (b) give the solutions obtained by the algorithm RVEA~\cite{Cheng2016} on a problem with a regular Pareto front (DTLZ1~\cite{Deb2005a} having a triangle Pareto front), and another problem with an irregular Pareto front (IDTLZ1~\cite{Deb2014} having an inverted triangle Pareto front), respectively. 
As can be seen from the figure, RVEA performs very well on the DTLZ1 (Figure~\ref{fig:sec2}(a)) but performs poorly on IDTLZ1 (Figure~\ref{fig:sec2}(b)).

In order to tackle this issue, researchers have proposed weight adaptation methods that adjust the weights to approximate the shape of the Pareto front during the evolutionary process. Table~\ref{tab:Sec2} (lower part) lists several representative weight adaptation algorithms, including MOEA/D-AM2M~\cite{Qi2014}, RVEA*~\cite{Cheng2016}, iRVEA~\cite{Liu2019b}, MOEA/D-LTD~\cite{Wu2018}, DEA-GNG~\cite{Liu2019}, RVEA-iGNG~\cite{Liu2020}, AdaW~\cite{Li2020}, MOEA/D-AM2M~\cite{Liu2017}, A-NSGA-III~\cite{Deb2014}, and g-DBEA~\cite{Asafuddoula2017}. 
With adaptation of its weights, decomposition-based MOEAs can work much better on problems with irregular Pareto fronts.
For example, 
as shown in Figure~\ref{fig:sec2}(c), 
the algorithm RVEA-iGNG~\cite{Liu2020} can obtain a set of well-diversified solutions on IDTLZ1,
compared to the solution set obtained by RVEA in Figure~\ref{fig:sec2}(b).

However, adapting weights during the search process can lead to degradation in performance on problems with regular Pareto fronts. 
Adapting the weights means changing search directions, 
and it affects not only the convergence of the population but also the diversity since the adaptation is conducted based on the current population which may not be able to reflect the problem's true Pareto front.  
Figure~\ref{fig:sec2}(d) gives the results of the weight adaptation algorithm RVEA-iGNG on the regular problem DTLZ1. 
As shown, the obtained solution set performs worse than that obtained by RVEA (Figure~\ref{fig:sec2}(a)). \footnote{It is worth mentioning that, apart from these weight adaptation methods, there are other methods being explored to address MOPs with irregular Pareto fronts. For instance, the utilisation of multiple sets of weights~\cite{He2023} and the application of adaptive scalarising functions~\cite{Ishibuchi2009, Wang2016, Pang2022} have also demonstrated robust performance in dealing with MOPs with irregular Pareto fronts. However, these methods also show somewhat weaker performance on MOPs with regular Pareto fronts.}

In this paper, we make an attempt to address the above issues, with the aim of making decomposition-based MOEAs perform very well on irregular problems without compromising their good performance on regular ones. We do so by proposing a weight adaptation trigger mechanism, which will be detailed in the next section. 


In addition, 
it is worth mentioning that there are a couple of studies that attempted to balance the performance on regular and irregular Pareto fronts, though not being decomposition-based MOEAs, such as AR-MOEA~\cite{Tian2017}, VaEA~\cite{Xiang2016}, BCE-IBEA~\cite{Li2016}. 
They perform very well on both regular and irregular Pareto fronts. 
For comprehensive evaluation of the proposed algorithm, we will include them in our comparative experiments. The results are presented in the supplementary material. 

\section{The Proposed Weight Adaptation Trigger Mechanism (ATM) approach}
\subsection{Basic Idea}

\begin{figure}
\centering
\begin{subfigure}{0.23\textwidth}
  \centering
  \includegraphics[width=\linewidth]{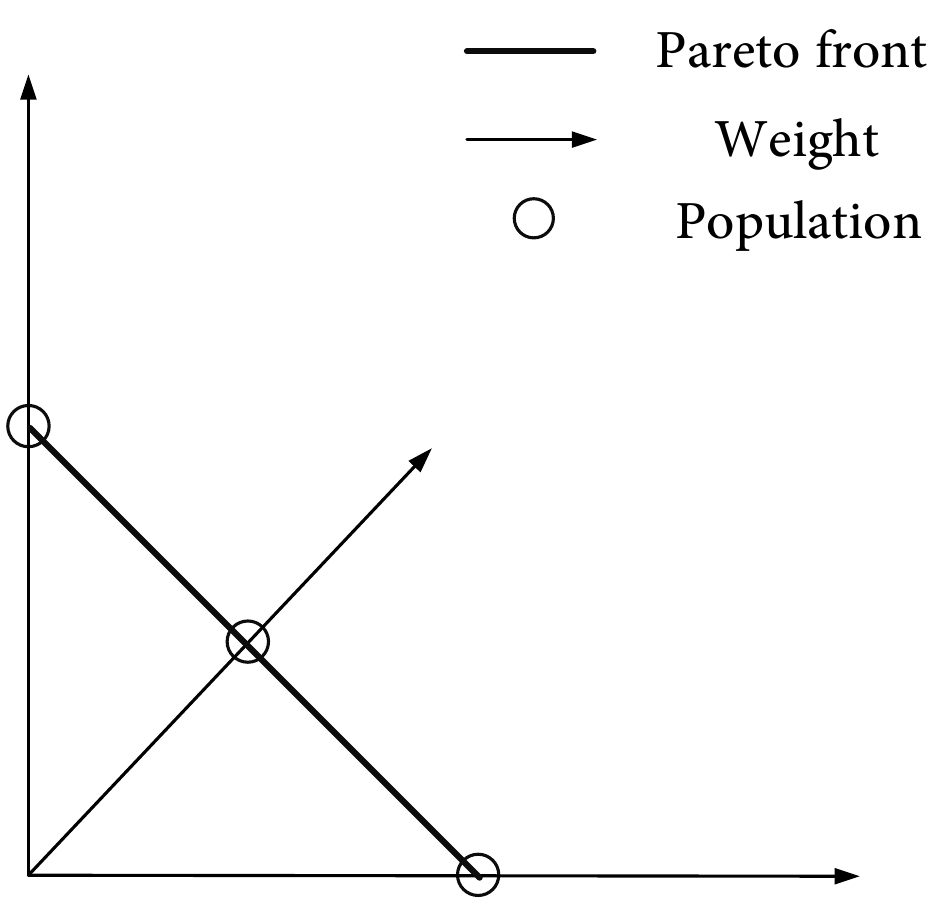}
  \caption{Evolutionary population}
  \label{fig:Idea1sub1}
\end{subfigure}%
\hspace{0cm}
\begin{subfigure}{0.21\textwidth}
  \centering
  \includegraphics[width=\linewidth]{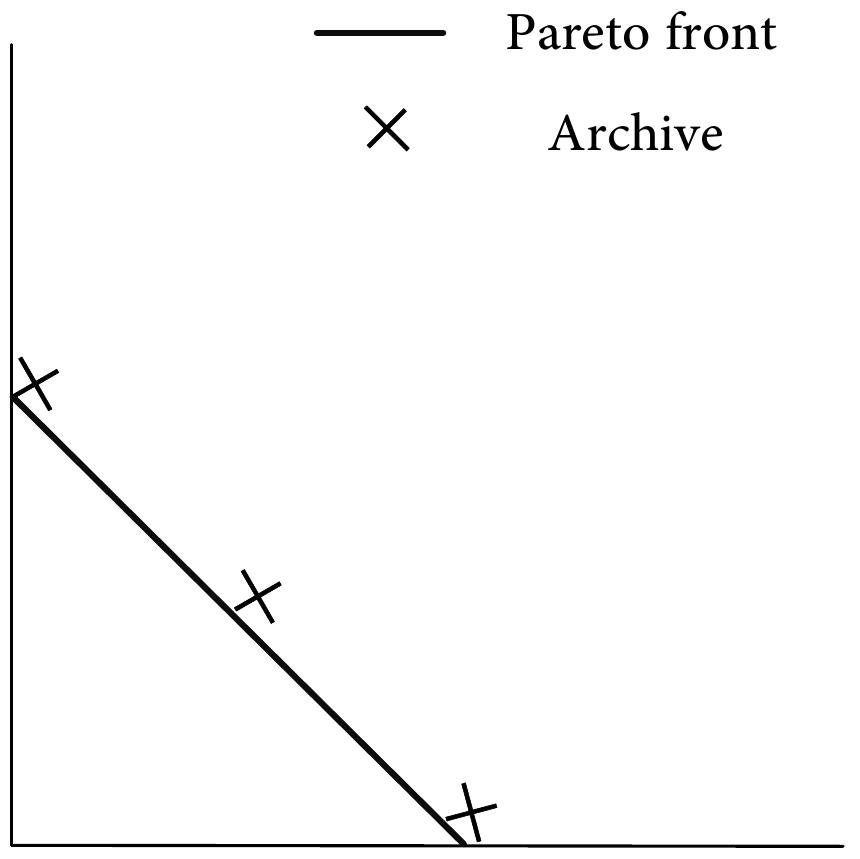}
  \caption{Archive}
  \label{fig:Idea1sub2}
\end{subfigure}
\caption{An illustration that the distribution of the stagnant population (i.e., the population close to the intersections between the weight directions and the Pareto front) is similar to that of the archive, so a regular Pareto front may be derived.}
\label{fig:Idea1}
\end{figure}

During the evolutionary process of decomposition-based MOEAs, the population is guided by the weights to move towards the intersections between the weight directions and the Pareto front~\cite{Zhang2007}. 
When the population is close
to these intersections, the evolution slows down gradually until full stagnation. 
In addition, during the evolutionary process, the non-dominated solutions generated in each generation can be stored in the archive. If the archive is properly maintained (by Pareto dominance and crowdedness-based criteria), its solutions may represent well-spread solutions found so far~\cite{Li2023}. 

\begin{figure}
\centering
\begin{subfigure}{0.22\textwidth}
  \centering
  \includegraphics[width=\linewidth]{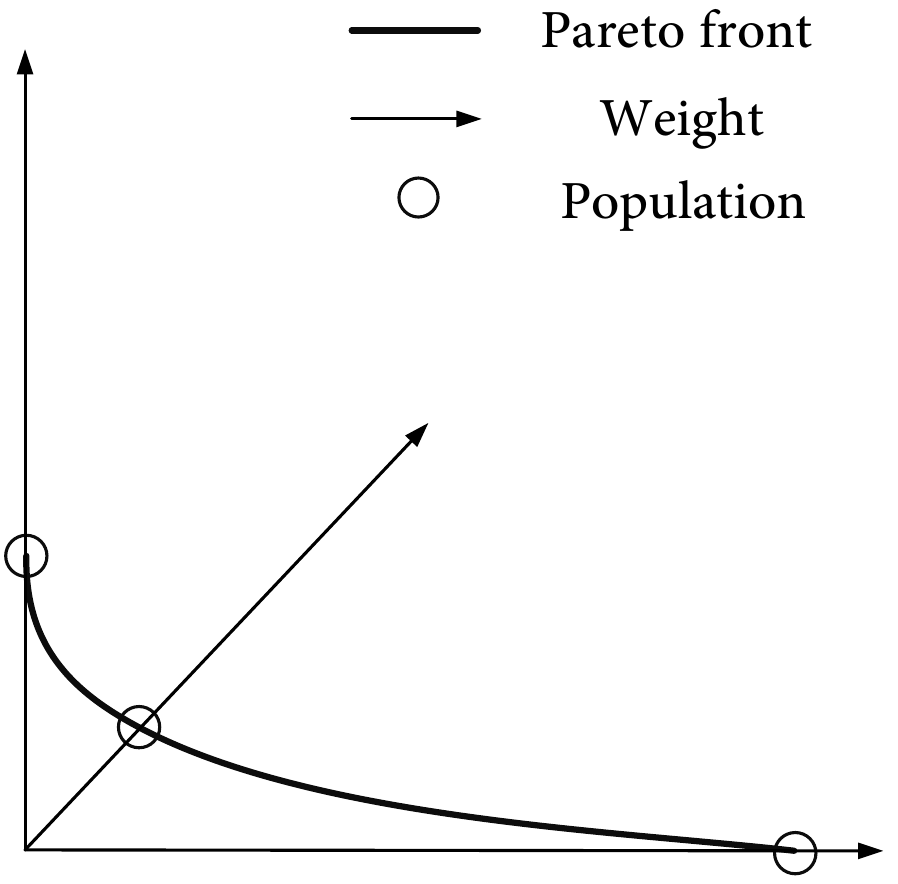}
  \caption{Evolutionary population}
  \label{fig:Idea2sub1}
\end{subfigure}
\hspace{0cm}
\begin{subfigure}{0.22\textwidth}
  \centering
  \includegraphics[width=\linewidth]{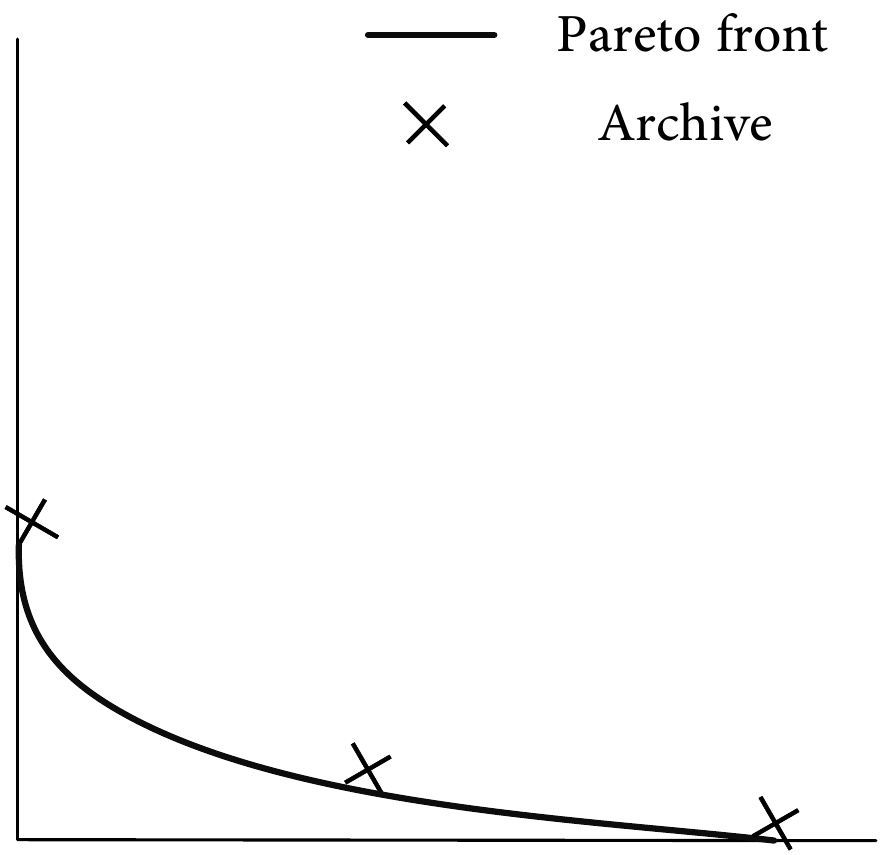}
  \caption{Archive}
  \label{fig:Idea2sub2}
\end{subfigure}
\caption{An illustration that the distribution of the stagnant population is not very similar to that of the archive (as the population is distributed not very uniformly), so an irregular Pareto front may be derived.}
\label{fig:Idea2}
\end{figure}

On the basis of the above two observations, one may be able to know if the problem's Pareto front is regular or not. That is, when the evolution stagnates, one may make a comparison between the distribution of the solutions in the archive and that of the current stagnant population. 
If their distributions are similar (or consistent), then it is likely a regular Pareto front (i.e., the simplex-shaped weight distribution can well represent the Pareto front); if not, then likely an irregular Pareto front (i.e., the Pareto front may not be of a simplex-like shape). 
Figure~\ref{fig:Idea1} and Figure~\ref{fig:Idea2} illustrate these two situations. 
In Figure~\ref{fig:Idea1}, when the evolution stagnates (i.e., reaching
the intersections between the weight directions and the Pareto front), 
one can see that the distribution of the population is similar to that of the archive, thus deriving a regular Pareto front. 
In contrast, 
in Figure~\ref{fig:Idea2}, when the evolution stagnates, one can see that the distribution of the population is not very similar to that of the archive (i.e., the population is distributed not very uniformly), thus deriving an irregular Pareto front. 

The above outlines the basic idea of our ATM-MOEA/D. However, implementing it successfully requires addressing several critical issues.
They are:
\begin{itemize}
    \item How to maintain an archive whose solutions are distributed uniformly, thus reflecting the Pareto front well?
    \item How to detect when evolution stagnates?
    \item How to define the consistency between the population and the archive, and consequently determine whether weight adaptation is necessary?
    \item How to adapt the weights, if weight adaptation is deemed necessary?
\end{itemize}

In the following sections, we first present the general framework of our proposed algorithm ATM-MOEA/D and then proceed to describe how we address the four issues outlined above.

\subsection{General Framework of ATM-MOEA/D}

\begin{figure}[tpb]
	\begin{center}
		\footnotesize
		\hspace{100mm}\includegraphics[scale=1]{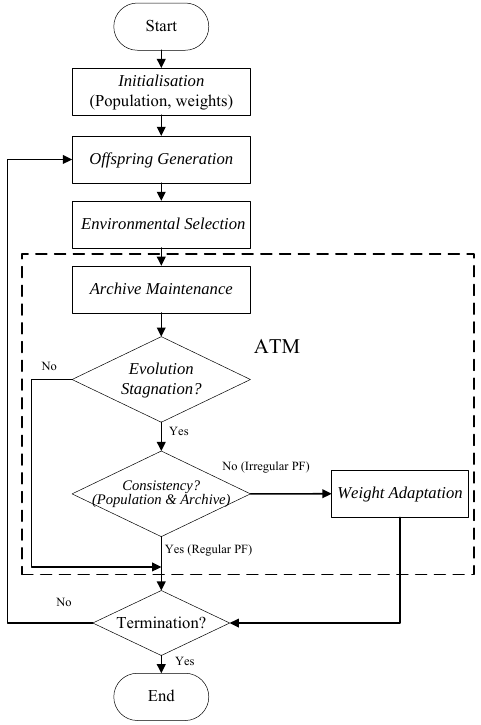}
	\end{center}
	\caption{The overall flowchart of ATM-MOEA/D.}
	\label{Framework3}
\end{figure}

Figure~\ref{Framework3} presents the overall flowchart of the proposed ATM-MOEA/D algorithm. 
It follows the basic framework of the original MOEA/D algorithm~\cite{Zhang2007}. 
That is, we borrowed components of \textit{Initialisation} (to initialise the population and weights), \textit{Offspring Generation} and \textit{Environmental Selection}. 
After \textit{Environmental Selection}, we introduced four additional components: \textit{Archive Maintenance}, \textit{Stagnation Detection} (to detect if the evolution stagnates or not), \textit{Consistency Detection} (to detect if the population and archive are consistent or not), and \textit{Weight Adaptation}. 
These four components, enclosed within a dotted line in the figure, collectively form the Weight Adaptation Trigger Mechanism (ATM), representing the primary contribution of our work.

ATM begins with maintaining a well-distributed archive that accurately represents the shape of the Pareto front (\textit{Archive Maintenance}). Then, it checks for evolution stagnation (\textit{Stagnation Detection}). If there is no stagnation, the algorithm continues without entering \textit{Consistency Detection}. 
However, if stagnation is detected, the consistency between the distribution of the population and the distribution of the archive is examined (\textit{Consistency Detection}). If they are consistent, indicating a regular Pareto front, 
the algorithm proceeds to the next iteration without \textit{Weight Adaptation}. 
If they are inconsistent, indicating an irregular Pareto front, \textit{Weight Adaptation} is activated.


\subsection{Archive Maintenance}
\label{Archive Maintenance}
A well-distributed archive plays a crucial role in ATM-MOEA/D, as it affects the algorithm's ability to accurately reflect the potential shape and characteristics of the problem's Pareto front.
During the archive maintenance process in ATM-MOEA/D, non-dominated solutions produced during the evolutionary process are stored in an archive with a pre-set capacity.
When the number of solutions in the archive exceeds its capacity, a subset of evenly distributed solutions are selected.
Inspired by~\cite{Wang2022}, we here present a subset selection method. 
Specifically, 
we first partition the current solutions in the archive into two sets: an main set and a backup set, 
where the size of the main set is set to the pre-defined capacity of the archive. 
We then insert the solutions from the backup set into the main set one by one, while simultaneously using a potential function~\footnote{The potential energy function is utilised as the objective function, which places significant emphasis on diversity while also having low computational complexity~\cite{Wang2022}. Recently, researchers in the evolutionary multi-objective optimisation field have incorporated similar ideas to enhance solution diversity~\cite{Wang2022, Falcon2019, Gomez2017}.} to remove a solution in order to enhance diversity among the remaining solutions in the first set.

Algorithm~\ref{Archive maintenance} presents further details on the archive maintenance method 
In each generation, non-dominated solutions generated during the evolutionary process are stored in the archive (line 1). If the number of solutions in the archive exceeds the capacity $N_A$ (line 2), 
the solutions in the archive are divided into the main set $A_M$ and the backup set $A_B$.
It is worth noting that the solutions in the archive $A$ can originate from either the archive of the last generation or newly generated offspring $O$ in the current generation. 
In $A$, all solutions from the archive of the previous generation are copied into the main set $A_M$.
If the number of solutions in $A_M$ is less than $N_A$, 
we randomly select solutions from the newly generated offspring $O$ in $A$ and add them to $A_M$ (line 3). 
The remaining solutions in $A$ are added to the backup set $A_B$ (line 4).
Next, the solutions from the backup set $A_B$ are put into the main set $A_M$ one by one.
After adding a solution, we use the potential energy function 
to select an existing solution from $A_M$ to remove. 
The potential energy function is designed to maximise the potential energy of the remaining solutions to improve their uniformity (lines 5--9). 
Finally, the algorithm returns the updated main set $A_M$.

Mathematically, the potential energy function of a solution set $P$ is defined as

\begin{equation}
	\label{equation1}
	E(P) = \underset{p^i \in P}{\sum}\underset{p^j \in P/p^i}{\sum} U(p^i,p^j), 
\end{equation}
where $U(p^i,p^j)$ is the potential energy between two points $p^i$ and $p^j$, which is defined as 

\begin{equation}
	\label{equation2}
	U(p^i,p^j) = \frac{1}{\mid\mid p^i-p^j \mid\mid^s}, 
\end{equation}
where $\mid\mid.\mid\mid$ denotes the module and $s$ denotes a control parameter. Following the practice in~\cite{Wang2022}, we set $s = 2M$ in our study.   

It is worth noting that ATM-MOEA/D adopts a ``steady'' archive maintenance approach, which adds one solution to the archive while simultaneously removing another solution. In contrast to many other archive maintenance methods~\cite{Li2016, Liu2020, Li2020} where multiple solutions can be added or removed at once, this steady approach leads to a decrease in the number of solutions added to or removed from the archive over time. Consequently, the archive tends to be stable, allowing the algorithm to adjust as few weights as possible during the following weight adaptation. This is beneficial to the convergence of decomposition-based MOEAs~\cite{Giagkiozis2013, Cheng2016, Qi2014, Li2020}.

\begin{algorithm}[tb]
	\caption{Archive maintenance}
	\label{Archive maintenance}
	\small
	\SetAlgoLined
	\KwIn{Archive $A$, offspring generated by population $O$, the capacity of the Archive $N_A$.}
	$A \leftarrow NondominatedSection(A \cup O)$\;
	\If{$\mid A\mid > N_A$} 
	{
		$A_M \leftarrow MainSetGeneration(A)$\;
		$A_B \leftarrow A/A_M$\;
		\For{$\forall \:a_B \in A_B$}
		{
			$A_M \leftarrow A_M \cup a_B$\;
			$a_M \leftarrow \underset{a_M \in A_M}{arg\,min}\ E(A_M/a_M) $\;
			$A_M \leftarrow A_M / a_M$;
		}

	} 
	\KwOut{$A_M$}
\end{algorithm}

\subsection{Detection of Evolution Stagnation}
\label{Detection of the evolution stagnation}



In decomposition-based MOEAs, 
each solution in the population is matched to the corresponding weight in every generation and is guided to move towards the intersection between the weight direction and the Pareto front.
When all the solutions are close to their intersections, 
the matching may be stable (i.e., for any weight, the matched solution is unchanged),
indicating that there is no drastic change of the solutions' quality.


Bearing this in mind, Algorithm~\ref{Stagnation Detection} outlines the procedure of \emph{Stagnation Detection}. In the $t$-th generation, we first calculate the number of weights corresponding to each solution in $P$. This is denoted by $Num_t$, a list of the number of weights corresponding to solutions in $P$ (line 1). 
Next, we determine whether there is a change in $Num_t$ in the current generation from that in the previous generation (lines 2--6). 
If $Num_t$ remains unchanged for $fre$ consecutive generations (where $fre$ is a parameter), we regard that the evolution has stagnated
(line 7).

\begin{algorithm}[tb]
	\caption{Stagnation Detection}
	\label{Stagnation Detection}
	\small
	\SetAlgoLined
	\KwIn{Solutions in the population at the $t$-th generation, $P = \{{p^t}_1, {p^t}_2,...{p^t}_i,...\}$, weights $W = \{w_1, w_2,...w_i,...\}$, the parameter $fre$.}
	
	  $Num_t \leftarrow  CorrespondingWeightNumber(P,W) $\tcp*{Calculate the corresponding weight number of each solution in $P$, where $Num_t = \{n_{{p^t}_1}, n_{{p^t}_2},...,n_{{p^t}_i}...\}$.}
	
	  \uIf{$Num_t = Num_{t-1}$} 
      {
    	$ count_t = 1 $ \tcp*{If the corresponding weights of all solutions in $P$ are unchanged, set the stagnation count to $1$.}
      }
      \Else
      {
    	$ count_t = 0 $ \tcp*{Otherwise, reset the stagnation count to $0$.}
      } 
     $Count \leftarrow \sum_{max\{0,t-fre+1\}}^{t} count_t$ \tcp*{If $Count = fre$, the evolution is stagnant because 
     the corresponding weight (or weights) of any solution in the population is unchanged in $fre$ consecutive generations.}
	\KwOut{$Count$}
\end{algorithm}

\subsection{Consistency Detection between the Population and Archive}
\label{Detection of the consistency between Population and Archive}

When the evolution stagnates, we know the population is close to the best possible status that it can achieve under the given weights. 
Therefore, it is time to decide if we need to change the weights. 
To do so, we check whether the distribution of the population is similar to the archive's distribution. 
If there does not exist a solution in the archive whose niche has no solution in the population, 
we regard that the distribution of the population and archive are similar.

Algorithm~\ref{ALG2} outlines the detailed procedure of \emph{Consistency Detection}.
First, we normalise all the solutions in the population and archive with respect to the minimum and maximum objective values of solutions in the archive (line 1). 
For each solution ${a^{\prime}}_i$ in the archive, we compute its distances to its closest solution in both the population (denoted by $d({a^{\prime}}_i)$) and the archive (denoted as $r({a^{\prime}}_i)$) (lines 2--5).
Then we calculate the niche size (or threshold) $r$ (line 6).
If the distance between all solutions in the archive and their closest solutions in the population is less than $r$, we regard the distribution of the archive and the distribution of the population to be similar (lines 8--10). Otherwise, we set $consistency = 0$ to indicate the inconsistency of the two sets (line 7).

\begin{algorithm}[tb]
	\caption{Consistency Detection}
	\label{ALG2}
	\small
	\SetAlgoLined
 \KwIn{Population $P$, Archive $A$}
	 $(P^{\prime},A^{\prime}) \leftarrow  Normalisation(P,A)$\;
      
	\For{$a^{\prime} \in A^{\prime}$}
	{
		$ r(a^{\prime}) \leftarrow MinimumDistanceCal(a^{\prime},A^{\prime}/a^{\prime})$ \;
		$ d(a^{\prime}) \leftarrow MinimumDistanceCal(a^{\prime},P^{\prime})$ \;
	}
  
    $r = \sqrt{m} \times \underset{a^{\prime} \in A^{\prime}}{Median} ~r(a^{\prime}) $ \;
    $ consistency = 0 $\;
    \If{$\forall a^{\prime} \in A^{\prime}, d(a^{\prime}) < r $ } 
    	{
    		$ consistency = 1 $\;
    	}
	\KwOut{$consistency$}
\end{algorithm}

%

\subsection{Weight Adaptation}

ATM-MOEA/D adopts the \emph{Weight Adaptation} technique from AdaW, which adapts weights through an ``add-delete'' approach~\cite{Li2020}.
However, when deleting weights, our method differs from AdaW in the case that each solution corresponds to only one weight (for the detail of this case, please refer to~\cite{Li2020}). In this case, 
we need to delete some weights and their corresponding solutions (in order to ensure that there are only population-size weights for the population). 
In AdaW, the algorithm defines the crowding degree of each solution and iteratively deletes the most crowded solutions based on their crowding degree. In our ATM-MOEA/D, we use the diversity maintenance method described in Section~\ref{Archive Maintenance} to select solutions that are evenly distributed from the current population. This method is more beneficial for improving the diversity of the population, which will be verified in our experiments. 

\section{Experimental design and results}
In this section, we begin with by introducing the experimental design, 
including a brief description of the peer algorithms for comparison, test problems, quality indicators, and parameter setting.
Then based on that, we verify the proposed ATM-MOEA/D\footnote{The source code for ATM-MOEA/D is available for download using the following link: \url{https://hanx63520.wixsite.com/atm-moea}.}. 

\subsection{Experimental Design}
To verify the advantages of ATM-MOEA/D on problems with both regular and irregular Pareto fronts, 
we compare it with seven peer algorithms\footnote{The codes of all the peer algorithms were from PlatEMO~\cite{Tian2017b}.}: MOEA/D~\cite{Zhang2007}, DEA-GNG~\cite{Liu2019}, RVEA-iGNG~\cite{Liu2020}, AdaW~\cite{Li2020}, VaEA~\cite{Xiang2016}, AR-MOEA~\cite{Tian2017}, and BCE-IBEA~\cite{Li2016}. 
MOEA/D is a well-known fixed-weight
decomposition-based algorithm that excels in solving problems with regular Pareto fronts,
thus serving as a baseline in our experimental study.
DEA-GNG~\cite{Liu2019}, RVEA-iGNG~\cite{Liu2020}, and AdaW~\cite{Li2020} are weight adaptation algorithms that demonstrate good performance on irregular problems, 
so being able to verify how our algorithm performs on irregular problems.
VaEA~\cite{Xiang2016}, AR-MOEA~\cite{Tian2017}, and BCE-IBEA~\cite{Li2016} are MOEAs not based on the idea of decomposition, 
but they exhibit very promising performance on both regular and irregular problems, 
so being included in our study as well (the comparison results are given in the supplementary material). 
Next, we provide a brief introduction to these algorithms.

\begin{itemize}
	\item MOEA/D~\cite{Zhang2009} initialises a number of uniformly-distributed weights to decompose a multi-objective problem into a number of single-objective sub-problems and then optimises them cooperatively. Throughout the evolutionary process, MOEA/D does not adjust the weights. As mentioned earlier, fixed-weight decomposition-based algorithms perform well on problems with regular Pareto fronts.
	
	
	
	\item DEA-GNG~\cite{Liu2019} adapt weights during the evolutionary process. DEA-GNG learns topological structures of Pareto fronts through a growing neural gas (GNG). Both the weights (reference vectors) and the scalarising functions are adapted based on the learned topological nodes. 

	\item RVEA-iGNG~\cite{Liu2020} is a weight-adapting version of RVEA~\cite{Cheng2016}. It learns the topological structure of the Pareto front through an improved growing neural gas (iGNG) and then adapts the weights (reference vectors) at each generation.

    \item AdaW~\cite{Liu2020} is a weight-adaptive algorithm that periodically updates weights by comparing the current evolutionary population with a well-maintained archive set. An advantage of AdaW is its ability to obtain a uniformly distributed set of solutions in problems with various Pareto front shapes.
    
    \item VaEA~\cite{Xiang2016} utilises the maximum vector angle first principle to select the solutions, ensuring their diversity on different Pareto front shapes. Moreover, by applying the worse-elimination principle, VaEA allows for the conditional replacement of poorer solutions based on their convergence, thereby enhancing the selection pressure.

	\item AR-MOEA~\cite{Tian2017} is an indicator-based MOEA. It adjusts a set of weights (reference points) by introducing an adaptation method based on the indicator contributions of candidate solutions in an external archive. These reference points guide the evolution of the population. AR-MOEA is known for its versatility in solving problems with various types of Pareto fronts.

    \item BCE-IBEA~\cite{Li2016} is an MOEA that operates within the bi-criterion evolution (BCE) framework. This framework integrates both the Pareto criterion (PC) and non-Pareto criterion (NPC). The PC evolution and NPC evolution work together, exchanging information to facilitate their respective evolution. The PC evolution strategy compensates for the potential loss of population diversity in NPC evolution, particularly in problems with irregular Pareto fronts. 
\end{itemize}

We selected typical 36 test problems. 
These test problems are taken from commonly used test suites in the field of multi-objective optimisation, including DTLZ~\cite{Deb2005a}, SDTLZ~\cite{Deb2014}, IDTLZ~\cite{Jain2014}, ZDT~\cite{Zitzler2000}, VNT~\cite{Veldhuizen1999}, FON~\cite{Fonseca1995}, SCH~\cite{Schaffer1985}, MaF~\cite{Cheng2017}, and IMOP~\cite{Tian2019}. 
We divided these problems into two categories: problems with regular Pareto fronts and problems with irregular Pareto fronts.
The test problems with regular Pareto fronts include the 2-objective DTLZ1, the 2-objective DTLZ2, the 2-objective DTLZ4, the 3-objective DTLZ1, the 3-objective DTLZ2, the 3-objective DTLZ4, the 5-objective DTLZ1, the 5-objective DTLZ2, ZDT2, ZDT6.
For the remaining problems with irregular Pareto fronts, we classified them into several categories according to their different characteristics. They include 1) problems with an inverted simplex-like Pareto front, such as the 3-objective inverted DTLZ1 (IDTLZ1), the 3-objective inverted DTLZ2 (IDTLZ2), the 10-objective inverted DTLZ1 (IDTLZ1) and the 3-objective MaF1; 2) problems with a highly nonlinear Pareto front such as SCH1, FON1, the 3-objective convex DTLZ2 (CDTLZ2), the 3-objective MaF2, the 3-objective MaF3, IMOP1, and IMOP2; 3) problems with a disconnect Pareto front such as ZDT3, the 3-objective DTLZ7, the 3-objective MaF7, IMOP3, and IMOP5; 4) problems with a degenerate Pareto front such as the 3-objective DTLZ5, the 3-objective MaF6, and IMOP4; 5) problems with a scaled Pareto front such as the 3-objective scaled DTLZ1 (SDTLZ1), the 3-objective scaled DTLZ2 (SDTLZ2), the 3-objective MaF5; 6) problems with several of the above characteristics such as SCH2 (which has a disconnect and scaled Pareto front), VNT2 (which has a degenerate and scaled Pareto front), the 3-objective MaF4 (which has an inverted and scaled Pareto front) and IMOP6 (which has a disconnect and degenerate Pareto front). The configuration of all these 36 test problems is the same as those in the original papers.

The inverted generational distance (IGD) \cite{Bosman2003} and Hypervolume (HV) \cite{Zitzler1999} quality indicators are adopted to assess the performance of the algorithms. 
Both the IGD and HV indicators evaluate the comprehensive quality of solution sets in terms of convergence and diversity.
The reason we choose IGD, coupled with HV, is its distinct behaviour, which can make a good complement to HV, provided that a set of uniform and dense points that well represents the Pareto front is available\cite{Li2020b}, as it is here. Note that for problems with a scaled Pareto front, the solutions and the reference set (in IGD) should be normalised based on the true Pareto front to calculate IGD.

To calculate HV, reference points that are dominated by the reference sets are required. In accordance with the practice in \cite{Li2020}, we set the reference points for the following problems: DTLZ1, DTLZ2, DTLZ5, DTLZ7, IDTLZ1, IDTLZ2, CDTLZ2, SDTLZ1, SDTLZ2, SCH1, SCH2, FON1, ZDT3, and VNT2 to (1, 1, ..., 1), (2, 2, ..., 2), (2, 2, 2), (2, 2, 7), (1, 1, ..., 1), (2, 2, 2), (2, 2, 2), (0.55, 5.5, 55), (1.1, 11, 110), (5, 5), (2, 17), (2, 2), (2, 2), and (5, 16, 12), respectively. For the remaining problems, we used the following common settings: (1, 1, 1) for MaF1, MaF2, and MaF3, (2, 4, 8) for MaF4, (8, 4, 2) for MaF5, (0.8, 0.8, 1) for MaF6, (2, 2, 7) for MaF7, (1, 1) for IMOP1 and IMOP2, (1.5, 1) for IMOP3, (1, 1, 1) for IMOP4, and (0.5, 0.5, 1.5) for IMOP6. It is worth noting that the range of the Pareto front is taken into account when defining the reference point. Therefore, the solution sets do not need to be normalised when measuring the HV value of the scaled problems~\cite{Li2019}.

Additionally, to provide a visual representation of the search behaviour of the seven algorithms, we plotted their final solution sets on representative test problems for a single run. This run was selected based on the solution set that has the median of the IGD values out of all the runs.

All of the algorithms utilised real-valued variables. To generate offspring, we employed the widely used Simulated Binary Crossover (SBX)\cite{Deb1995} with a crossover probability of $p_c = 1.0$, and Polynomial Mutation (PM)\cite{Deb2001} with a mutation probability of $p_m = 1/d$, where $d$ represents the number of decision variables. The distribution parameter index was set to $\eta_c = 20$ for PM. These parameters were configured based on their original studies.

For decomposition-based MOEAs, 
the population size, which is the same as the number of weights, cannot be arbitrarily set. To obtain a set of uniformly distributed weights in a simplex, we fixed the population size to 100, 105, 220, and 220 for 2-, 3-, 5-, and 10-objective problems, respectively. Following the practice in~\cite{Liu2020}, we set the number of function evaluations to 50,000, 100,000, 150,000, and 150,000 for 2-, 3-, 5-, and 10-objective problems, respectively. We conducted 30 independent runs of each algorithm on each test problem.

The parameters of the peer algorithms were set according to their original papers.  
For MOEA/D, the neighbourhood size was set to $10\%$ of the population size.
For DEA-GNG, $\epsilon$ was set to $ 0.05\pi $ for 2- and 3-objective problems, $ 0.15\pi $ for 5-objective problems, and $ 0.2\pi $ for 10 - and 15-objective problems. The archive size was set to $ m \times N $, where $ m $ is the number of objectives and $ N $ is the population size.
In RVEA-iGNG, the parameters of the iGNG were set to $ \varepsilon_b = 0.2 $, $ \varepsilon_n = 0.006 $, $ \alpha = 0.5 $, $ \alpha_{max} = 50 $, $ d = 0.995 $, and $ \lambda = 50 $. The frequency of adjustment of the reference vectors (weights) was set to $ f_r = 0.1 $.
In AdaW, the maximum capacity of the archive was set to $ 2N $. 
The weights were adapted every $5\%$ of the total generations, and adaptation was not allowed during the last $10\%$ of generations.
In our ATM-MOEA/D, the number of the generations for determining the evolutionary stagnation $fre$ was set to $5\%$ of the total generations. The niche size $r$ (for consistency detection between the archive and population) is set to the product of $\sqrt{m} $ and the median distance between each solution in the archive and its nearest solution in the archive, where $ m $ is the number of objectives. The size of the archive is set to $2N$.

\begin{table*}[ht]
	\begin{footnotesize}
		\begin{threeparttable}
			\caption{IGD results (mean and SD) of the five algorithms.}
			\label{IGD-ATM-MOEA/D}
			\centering
			\begin{tabular}{ccccccc}
				\hline
				\specialrule{0em}{1pt}{1pt}
				Pareto front &Problem & MOEA/D & DEA-GNG & RVEA-iGNG & AdaW & ATM-MOEA/D\\
				\hline
				\specialrule{0em}{1pt}{1pt}
				\multirow{10}[0]{*}{Regular}&DTLZ1-2 & 1.836E-03(6.1E-05)$^\dagger$ & 2.173E-03(1.2E-04)$^\dagger$ & 1.979E-03(2.7E-05)$^\dagger$ & 1.888E-03(3.9E-05)$^\dagger$ & \textbf{1.798E-03(2.8E-05)}\\
				\specialrule{0em}{1pt}{1pt}
				&DTLZ2-2 & \textbf{4.133E-03(2.1E-08)}$^\dagger$ & 4.592E-03(1.3E-04)$^\dagger$ & 4.374E-03(4.8E-05)$^\dagger$ & 4.142E-03(2.2E-05)$^\dagger$ & 4.133E-03(3.8E-08)\\
				\specialrule{0em}{1pt}{1pt}
				&DTLZ4-2 & 1.772E-01(3.2E-01)$^\dagger$ & 1.530E-01(3.0E-01)$^\dagger$ & 2.762E-01(3.6E-01)$^\dagger$ & 2.886E-02(1.4E-01) & \textbf{4.135E-03(1.2E-05)}\\
				\specialrule{0em}{1pt}{1pt}
				&DTLZ1-3 & 1.872E-02(2.6E-05) & 5.609E-02(4.5E-02)$^\dagger$ & 1.947E-02(3.8E-04)$^\dagger$ & 1.919E-02(2.3E-04)$^\dagger$ & 1.873E-02(1.0E-04)\\
				\specialrule{0em}{1pt}{1pt}
				&DTLZ2-3 & 5.015E-02(3.8E-07) & 5.232E-02(1.9E-03)$^\dagger$ & 5.339E-02(1.2E-03)$^\dagger$ & 5.118E-02(7.0E-04)$^\dagger$ & 5.025E-02(3.3E-04)\\
				\specialrule{0em}{1pt}{1pt}
				&DTLZ4-3 & 3.525E-01(3.6E-01)$^\dagger$ & 5.236E-02(1.8E-03)$^\dagger$ & 1.521E-01(3.0E-01)$^\dagger$ & 6.417E-02(7.1E-02)$^\dagger$ & \textbf{5.031E-02(4.7E-04)}\\
				\specialrule{0em}{1pt}{1pt}
				&DTLZ1-5 & 5.409E-02(3.4E-03)$^\dagger$ & 1.964E-01(7.3E-02)$^\dagger$ & 4.932E-02(4.9E-04) & 5.096E-02(4.6E-04)$^\dagger$ & 5.205E-02(1.7E-04)\\
				\specialrule{0em}{1pt}{1pt}
				&DTLZ2-5 & 1.347E-01(4.6E-03)$^\dagger$ & 1.374E-01(6.4E-03)$^\dagger$ & 1.444E-01(3.2E-03)$^\dagger$ & 1.422E-01(5.3E-03)$^\dagger$ & \textbf{1.332E-01(1.1E-04)}\\
				\specialrule{0em}{1pt}{1pt}
				&ZDT2 & \textbf{3.807E-03(4.6E-07)}$^\dagger$ & 4.344E-03(1.6E-04)$^\dagger$ & 4.154E-03(8.5E-05)$^\dagger$ & 3.895E-03(3.4E-05)$^\dagger$ & 3.848E-03(6.1E-05)\\
				\specialrule{0em}{1pt}{1pt}
				&ZDT6 & 3.111E-03(5.3E-06) & 3.710E-03(8.5E-04)$^\dagger$ & 3.273E-03(5.6E-05)$^\dagger$ & 3.141E-03(4.0E-05)$^\dagger$ & 3.112E-03(1.1E-05)\\
				\hline
				\specialrule{0em}{1pt}{1pt}
				\multirow{36}[0]{*}{Irregular}&DTLZ5-3 & 3.130E-02(1.5E-05)$^\dagger$ & 5.065E-03(2.3E-04)$^\dagger$ & 4.357E-03(1.0E-04)$^\dagger$ & 4.048E-03(4.4E-05)$^\dagger$ & \textbf{4.015E-03(3.9E-05)}\\
				\specialrule{0em}{1pt}{1pt}
				&DTLZ7-3 & 1.544E-01(1.2E-01)$^\dagger$ & 6.703E-02(5.4E-02)$^\dagger$ & 8.468E-02(9.1E-02)$^\dagger$ & 5.298E-02(7.8E-04) & 5.356E-02(6.2E-04)\\
				\specialrule{0em}{1pt}{1pt}
				&CDTLZ2-3 & 1.089E-01(2.8E-04)$^\dagger$ & 5.496E-02(1.4E-02)$^\dagger$ & 3.443E-02(2.2E-03)$^\dagger$ & 2.830E-02(8.6E-04) & 2.848E-02(8.6E-04)\\
				\specialrule{0em}{1pt}{1pt}
				&IDTLZ1-3 & 2.962E-02(1.8E-05)$^\dagger$ & 2.733E-02(9.4E-03)$^\dagger$ & 2.006E-02(3.7E-04)$^\dagger$ & 8.661E-01(2.7E-04)$^\dagger$ & \textbf{1.931E-02(2.2E-04)}\\
				\specialrule{0em}{1pt}{1pt}
				&IDTLZ2-3 & 7.255E-02(1.2E-03)$^\dagger$ & 7.944E-02(1.4E-02)$^\dagger$ & 5.699E-02(5.5E-03)$^\dagger$ & 5.072E-02(6.0E-04) & 5.085E-02(4.9E-04)\\
				\specialrule{0em}{1pt}{1pt}
				&ZDT3 & 1.783E-02(9.7E-03)$^\dagger$ & 5.177E-03(3.1E-04)$^\dagger$ & 6.870E-03(7.4E-03)$^\dagger$ & 4.666E-03(9.7E-05) & \textbf{4.649E-03(7.0E-05)}\\
				\specialrule{0em}{1pt}{1pt}
				&FON1 & 4.939E-03(1.7E-04)$^\dagger$ & 4.979E-03(1.3E-04)$^\dagger$ & 4.817E-03(8.4E-05)$^\dagger$ & 4.737E-03(8.3E-05)$^\dagger$ & \textbf{4.585E-03(3.6E-06)}\\
				\specialrule{0em}{1pt}{1pt}
				&SCH1 & 1.801E-01(3.1E-03)$^\dagger$ & 1.909E-02(8.0E-04)$^\dagger$ & 1.947E-02(4.8E-04)$^\dagger$ & 1.714E-02(1.9E-04) & 1.720E-02(2.0E-04)\\
				\specialrule{0em}{1pt}{1pt}
				&SCH2 & 5.677E+00(5.5E-05)$^\dagger$ & 1.856E+00(4.1E-03)$^\dagger$ & 4.414E+00(1.8E+00)$^\dagger$ & 2.136E-02(4.0E-04) & \textbf{2.134E-02(2.1E-04)}\\
				\specialrule{0em}{1pt}{1pt}
				&SDTLZ1-3 & 1.051E+01(7.5E-03)$^\dagger$ & 3.951E+00(4.8E+00)$^\dagger$ & \textbf{5.087E-01(1.6E-02)}$^\dagger$ & 6.231E-01(3.1E-02) & \textbf{6.187E-01(3.1E-02)}\\
				\specialrule{0em}{1pt}{1pt}
				&SDTLZ2-3 & 1.721E+01(3.2E-03)$^\dagger$ & 1.426E+00(6.6E-02)$^\dagger$ & \textbf{1.157E+00(3.1E-02)}$^\dagger$ & 1.236E+00(4.5E-02) & 1.237E+00(4.1E-02)\\
				\specialrule{0em}{1pt}{1pt}
				&VNT2 & 5.993E-02(6.4E-05)$^\dagger$ & 1.274E-02(8.9E-04)$^\dagger$ & \textbf{1.085E-02(4.0E-04)}$^\dagger$ & 1.122E-02(2.4E-04)$^\dagger$ & 1.106E-02(1.7E-04)\\
				\specialrule{0em}{1pt}{1pt}
				&IDTLZ1-10 & 2.680E-01(3.6E-03)$^\dagger$ & 1.226E-01(1.2E-02)$^\dagger$ & 1.401E-01(1.4E-02)$^\dagger$ & 1.132E-01(3.0E-02)$^\dagger$ & \textbf{1.074E-01(8.4E-03)}\\
				\specialrule{0em}{1pt}{1pt}
				&MaF1-3 & 6.067E-02(4.6E-06)$^\dagger$ & 4.063E-02(8.8E-04)$^\dagger$ & 4.075E-02(4.3E-04)$^\dagger$ & 3.931E-02(1.8E-04) & 3.931E-02(1.7E-04)\\
				\specialrule{0em}{1pt}{1pt}
				&MaF2-3 & 3.851E-02(6.0E-05)$^\dagger$ & 2.845E-02(6.3E-04)$^\dagger$ & 2.811E-02(2.5E-04)$^\dagger$ & \textbf{2.803E-02(6.4E-04)}$^\dagger$ & 3.458E-02(6.5E-05)\\
				\specialrule{0em}{1pt}{1pt}
				&MaF3-3 & 1.041E-01(2.9E-04)$^\dagger$ & 5.890E-02(1.6E-02)$^\dagger$ & \textbf{3.198E-02(5.4E-04)}$^\dagger$ & 3.472E-02(2.1E-03) & 3.432E-02(2.5E-03)\\
				\specialrule{0em}{1pt}{1pt}
				&MaF4-3 & 5.501E-01(2.0E-03)$^\dagger$ & 3.468E-01(9.7E-02)$^\dagger$ & 2.489E-01(8.2E-03)$^\dagger$ & 2.463E-01(2.0E-02) & 2.538E-01(6.3E-02)\\
				\specialrule{0em}{1pt}{1pt}
				&MaF5-3 & 1.826E+00(1.8E+00)$^\dagger$ & 4.004E-01(8.5E-01)$^\dagger$ & 3.997E-01(8.5E-01)$^\dagger$ & 2.781E-01(2.3E-01) & 2.790E-01(2.3E-01)\\
				\specialrule{0em}{1pt}{1pt}
				&MaF6-3 & 1.241E-02(3.7E-06)$^\dagger$ & 4.996E-03(2.4E-04)$^\dagger$ & 4.343E-03(1.1E-04)$^\dagger$ & 4.041E-03(3.2E-05)$^\dagger$ & \textbf{3.997E-03(3.4E-05)}\\
				\specialrule{0em}{1pt}{1pt}
				&MaF7-3 & 2.282E-01(6.5E-02)$^\dagger$ & 9.890E-02(1.1E-01)$^\dagger$ & 8.493E-02(9.1E-02)$^\dagger$ & \textbf{5.319E-02(7.8E-04)}$^\dagger$ & 5.401E-02(8.3E-04)\\
				\specialrule{0em}{1pt}{1pt}
				&IMOP1 & 1.012E-01(8.7E-03)$^\dagger$ & 1.660E-02(3.3E-03)$^\dagger$ & 6.287E-02(1.8E-02)$^\dagger$ & 5.743E-03(1.5E-03) & \textbf{5.338E-03(3.8E-04)}\\
				\specialrule{0em}{1pt}{1pt}
				&IMOP2 & 6.185E-01(3.3E-02)$^\dagger$ & 9.647E-03(3.3E-03) & 9.918E-03(2.8E-03) & 1.448E-02(7.3E-03)$^\dagger$ & 1.175E-02(5.0E-03)\\
				\specialrule{0em}{1pt}{1pt}
				&IMOP3 & 2.036E-02(3.0E-02)$^\dagger$ & 7.297E-03(2.6E-03)$^\dagger$ & 5.239E-03(3.6E-03)$^\dagger$ & 5.770E-03(5.1E-03)$^\dagger$ & \textbf{3.743E-03(2.4E-04)}\\
				\specialrule{0em}{1pt}{1pt}
				&IMOP4 & 2.464E-02(2.8E-04)$^\dagger$ & 1.062E-02(1.7E-03)$^\dagger$ & 6.805E-03(1.0E-04)$^\dagger$ & 7.109E-03(4.0E-04)$^\dagger$ & \textbf{6.714E-03(1.1E-04)}\\
				\specialrule{0em}{1pt}{1pt}
				&IMOP5 & 5.082E-02(2.6E-04)$^\dagger$ & 3.953E-02(3.0E-03)$^\dagger$ & 3.281E-02(4.5E-04)$^\dagger$ & 3.216E-02(1.2E-03) & \textbf{3.187E-02(6.6E-04)}\\
				\specialrule{0em}{1pt}{1pt}
				&IMOP6 & 4.268E-02(6.0E-04)$^\dagger$ & 3.100E-02(3.0E-03)$^\dagger$ & 3.135E-02(3.3E-04)$^\dagger$ & 2.975E-02(4.4E-04) & 2.991E-02(2.2E-04)\\
				\hline
				\specialrule{0em}{1pt}{1pt}
				\multicolumn{2}{c}{$ +/=/- $ }& \textbf{$31/3/2$} &  \textbf{$35/1/0$} & \textbf{$30/1/5$} & \textbf{$17/16/3$} &  \\
				\hline
			\end{tabular}{}
			\begin{footnotesize}				
				$'\dagger'$ indicates that ATM-MOEA/D is of statistically significant difference from the corresponding peer algorithm at a 0.05 level by Wilcoxon's rank sum test. The best mean for each case is highlighted in boldface. The symbols $+, = $, and $- $ indicate that the results of ATM-MOEA/D are significantly better than, worse than, and equivalent to the corresponding peer algorithm.
			\end{footnotesize}
		\end{threeparttable}
	\end{footnotesize}
	
\end{table*}

\begin{table*}[ht]
	\begin{footnotesize}
		\begin{threeparttable}
			\caption{HV results (mean and SD) of the five algorithms.}
			\label{HV-ATM-MOEA/D}
			\centering
			\begin{tabular}{ccccccc}
				\hline
				\specialrule{0em}{1pt}{1pt}
				Pareto front & Problem & MOEA/D & DEA-GNG & RVEA-iGNG & AdaW & ATM-MOEA/D\\
				\hline
				\specialrule{0em}{1pt}{1pt}
				\multirow{10}[0]{*}{Regular} &DTLZ1-2 & 8.736E-01(1.4E-04)$^\dagger$ & 8.733E-01(1.7E-04)$^\dagger$ & 8.735E-01(9.1E-05)$^\dagger$ & 8.732E-01(6.8E-04)$^\dagger$ & \textbf{8.737E-01(1.5E-04)}\\
				\specialrule{0em}{1pt}{1pt}
				&DTLZ2-2 & 3.210E+00(2.7E-07)$^\dagger$ & \textbf{3.210E+00(1.4E-04)}$^\dagger$ & 3.210E+00(4.2E-04)$^\dagger$ & 3.210E+00(2.3E-03)$^\dagger$ & 3.210E+00(1.6E-07)\\
				\specialrule{0em}{1pt}{1pt}
				&DTLZ4-2 & 2.928E+00(5.2E-01)$^\dagger$ & 2.968E+00(4.9E-01)$^\dagger$ & 2.766E+00(5.9E-01) & 3.168E+00(2.2E-01)$^\dagger$ & \textbf{3.210E+00(7.0E-05)}\\
				\specialrule{0em}{1pt}{1pt}
				&DTLZ1-3 & 9.741E-01(4.0E-05) & 9.467E-01(1.8E-02)$^\dagger$ & 9.737E-01(1.2E-04)$^\dagger$ & 9.739E-01(1.6E-04)$^\dagger$ & \textbf{9.741E-01(3.1E-05)}\\
				\specialrule{0em}{1pt}{1pt}
				&DTLZ2-3 & 7.418E+00(4.3E-06) & 7.349E+00(2.6E-02)$^\dagger$ & 7.414E+00(3.5E-03)$^\dagger$ & 7.412E+00(6.9E-03)$^\dagger$ & 7.418E+00(2.7E-03)\\
				\specialrule{0em}{1pt}{1pt}
				&DTLZ4-3 & 6.605E+00(1.1E+00)$^\dagger$ & 7.372E+00(1.8E-02)$^\dagger$ & 7.070E+00(1.0E+00)$^\dagger$ & 7.380E+00(1.8E-01)$^\dagger$ & \textbf{7.418E+00(3.2E-03)}\\
				\specialrule{0em}{1pt}{1pt}
				&DTLZ1-5 & 9.988E-01(3.1E-04)$^\dagger$ & 8.420E-01(1.1E-01)$^\dagger$ & 9.989E-01(3.6E-05)$^\dagger$ & 9.990E-01(4.2E-05)$^\dagger$ & \textbf{9.990E-01(5.0E-06)}\\
				\specialrule{0em}{1pt}{1pt}
				&DTLZ2-5 & 3.169E+01(2.6E-02)$^\dagger$ & 3.085E+01(2.0E-01)$^\dagger$ & 3.168E+01(5.0E-03)$^\dagger$ & 3.169E+01(1.1E-02)$^\dagger$ & \textbf{3.170E+01(9.1E-05)}\\
				\specialrule{0em}{1pt}{1pt}
				&ZDT2 & \textbf{3.328E+00(3.1E-05)}$^\dagger$ & 3.328E+00(1.6E-04)$^\dagger$ & 3.328E+00(6.3E-04) & 3.328E+00(2.0E-03) & 3.328E+00(2.5E-03)\\
				\specialrule{0em}{1pt}{1pt}
				&ZDT6 & 3.041E+00(2.4E-04)$^\dagger$ & 3.040E+00(4.7E-03) & \textbf{3.042E+00(5.0E-05)}$^\dagger$ & 3.039E+00(4.2E-03)$^\dagger$ & 3.040E+00(3.8E-03)\\
				\hline
				\specialrule{0em}{1pt}{1pt}
				\multirow{36}[0]{*}{Irregular}&DTLZ5-3 & 6.042E+00(2.2E-05)$^\dagger$ & 6.103E+00(4.8E-04)$^\dagger$ & 6.103E+00(2.3E-04)$^\dagger$ & 6.103E+00(1.1E-04)$^\dagger$ & \textbf{6.105E+00(1.1E-04)}\\
				\specialrule{0em}{1pt}{1pt}
				&DTLZ7-3 & 1.313E+01(1.1E+00)$^\dagger$ & 1.307E+01(5.6E-01)$^\dagger$ & 1.317E+01(9.5E-01)$^\dagger$ & 1.349E+01(2.5E-02)$^\dagger$ & \textbf{1.351E+01(9.4E-03)}\\
				\specialrule{0em}{1pt}{1pt}
				&CDTLZ2-3 & 7.907E+00(3.6E-03)$^\dagger$ & 7.909E+00(2.2E-02)$^\dagger$ & 7.950E+00(9.4E-04)$^\dagger$ & \textbf{7.952E+00(1.8E-04)}$^\dagger$ & 7.951E+00(1.8E-04)\\
				\specialrule{0em}{1pt}{1pt}
				&IDTLZ1-3 & 6.678E-01(6.3E-05)$^\dagger$ & 6.637E-01(1.5E-02)$^\dagger$ & 6.878E-01(4.6E-04) & 6.849E-01(3.4E-03)$^\dagger$ & 6.877E-01(9.1E-04)\\
				\specialrule{0em}{1pt}{1pt}
				&IDTLZ2-3 & 6.639E+00(9.0E-03)$^\dagger$ & 6.584E+00(4.5E-02)$^\dagger$ & 6.691E+00(1.6E-02)$^\dagger$ & 6.728E+00(3.3E-03) & \textbf{6.729E+00(3.7E-03)}\\
				\specialrule{0em}{1pt}{1pt}
				&ZDT3 & 4.731E+00(1.1E-01)$^\dagger$ & 4.812E+00(8.5E-04)$^\dagger$ & 4.791E+00(9.3E-02)$^\dagger$ & 4.814E+00(2.9E-03)$^\dagger$ & \textbf{4.815E+00(4.2E-05)}\\
				\specialrule{0em}{1pt}{1pt}
				&FON1 & 3.062E+00(7.1E-05)$^\dagger$ & 3.062E+00(1.3E-03)$^\dagger$ & \textbf{3.063E+00(5.7E-05)}$^\dagger$ & 3.058E+00(5.6E-03) & 3.062E+00(3.3E-05)\\
				\specialrule{0em}{1pt}{1pt}
				&SCH1 & 2.197E+01(7.6E-03)$^\dagger$ & 2.227E+01(2.8E-03)$^\dagger$ & \textbf{2.228E+01(7.9E-04)}$^\dagger$ & 2.227E+01(9.9E-04)$^\dagger$ & 2.227E+01(1.1E-03)\\
				\specialrule{0em}{1pt}{1pt}
				&SCH2 & 3.365E+01(5.4E-04)$^\dagger$ & 3.462E+01(5.4E-02)$^\dagger$ & 3.396E+01(4.3E-01)$^\dagger$ & 3.824E+01(3.3E-02) & \textbf{3.825E+01(2.1E-03)}\\
				\specialrule{0em}{1pt}{1pt}
				&SDTLZ1-3 & 9.687E+01(3.5E-02)$^\dagger$ & 1.265E+02(1.3E+01)$^\dagger$ & 1.392E+02(1.4E-01)$^\dagger$ & 1.403E+02(5.4E-02) & 1.403E+02(7.7E-02)\\
				\specialrule{0em}{1pt}{1pt}
				&SDTLZ2-3 & 4.858E+02(5.4E-02)$^\dagger$ & 7.333E+02(2.6E+00)$^\dagger$ & 7.410E+02(2.0E+00)$^\dagger$ & 7.484E+02(1.1E+00) & \textbf{7.489E+02(8.8E-01)}\\
				\specialrule{0em}{1pt}{1pt}
				&VNT2 & 1.609E+03(2.1E-02)$^\dagger$ & 1.647E+03(1.2E+00)$^\dagger$ & 1.647E+03(3.3E-01)$^\dagger$ & 1.648E+03(6.2E-03) & 1.648E+03(8.6E-03)\\
				\specialrule{0em}{1pt}{1pt}
				&IDTLZ1-10 & 3.188E-08(4.0E-09) & \textbf{4.520E-07(1.3E-07)}$^\dagger$ & 3.516E-07(2.3E-07)$^\dagger$ & 4.407E-07(5.5E-07) & 2.625E-07(3.8E-07)\\
				\specialrule{0em}{1pt}{1pt}
				&MaF1-3 & 1.050E-01(4.6E-06)$^\dagger$ & 1.300E-01(7.9E-04) & 1.280E-01(6.1E-04)$^\dagger$ & 1.299E-01(2.8E-04) & 1.299E-01(2.9E-04)\\
				\specialrule{0em}{1pt}{1pt}
				&MaF2-3 & 2.602E-01(9.1E-05)$^\dagger$ & 2.606E-01(1.5E-03)$^\dagger$ & \textbf{2.654E-01(3.6E-04)}$^\dagger$ & 2.652E-01(6.5E-04)$^\dagger$ & 2.621E-01(4.8E-04)\\
				\specialrule{0em}{1pt}{1pt}
				&MaF3-3 & 9.171E-01(1.1E-03)$^\dagger$ & 9.331E-01(9.8E-03)$^\dagger$ & 9.504E-01(6.2E-04) & 9.501E-01(9.6E-04) & 9.501E-01(1.2E-03)\\
				\specialrule{0em}{1pt}{1pt}
				&MaF4-3 & 2.630E+01(5.2E-02)$^\dagger$ & 2.883E+01(1.5E+00) & \textbf{2.955E+01(1.1E-01)}$^\dagger$ & 2.924E+01(3.8E-01) & 2.917E+01(6.8E-01)\\
				\specialrule{0em}{1pt}{1pt}
				&MaF5-3 & 1.445E+01(8.8E+00)$^\dagger$ & 2.540E+01(4.8E+00)$^\dagger$ & 2.577E+01(4.9E+00)$^\dagger$ & 2.637E+01(2.4E+00)$^\dagger$ & \textbf{2.638E+01(2.4E+00)}\\
				\specialrule{0em}{1pt}{1pt}
				&MaF6-3 & 7.988E-02(1.7E-06)$^\dagger$ & 8.284E-02(9.2E-05)$^\dagger$ & 8.296E-02(5.7E-05)$^\dagger$ & 8.301E-02(3.2E-05) & 8.301E-02(2.0E-05)\\
				\specialrule{0em}{1pt}{1pt}
				&MaF7-3 & 1.288E+01(1.0E+00)$^\dagger$ & 1.278E+01(1.0E+00)$^\dagger$ & 1.317E+01(9.5E-01)$^\dagger$ & 1.348E+01(3.2E-02)$^\dagger$ & \textbf{1.351E+01(5.5E-03)}\\
				\specialrule{0em}{1pt}{1pt}
				&IMOP1 & 9.822E-01(3.8E-04)$^\dagger$ & 9.850E-01(6.4E-05)$^\dagger$ & \textbf{9.850E-01(9.2E-05)}$^\dagger$ & 9.846E-01(6.3E-05) & 9.846E-01(6.3E-05)\\
				\specialrule{0em}{1pt}{1pt}
				&IMOP2 & 3.997E-04(4.1E-04)$^\dagger$ & 7.004E-02(5.1E-04)$^\dagger$ & \textbf{7.083E-02(7.6E-05)}$^\dagger$ & 7.065E-02(5.7E-04) & 7.070E-02(5.1E-04)\\
				\specialrule{0em}{1pt}{1pt}
				&IMOP3 & 1.104E+00(2.2E-02)$^\dagger$ & 1.113E+00(1.6E-03)$^\dagger$ & 1.115E+00(1.9E-03) & 1.113E+00(4.3E-03)$^\dagger$ & \textbf{1.115E+00(1.5E-03)}\\
				\specialrule{0em}{1pt}{1pt}
				&IMOP4 & 3.355E-01(2.5E-04)$^\dagger$ & 3.404E-01(1.8E-03)$^\dagger$ & 3.438E-01(3.0E-04)$^\dagger$ & 3.438E-01(3.6E-04)$^\dagger$ & \textbf{3.441E-01(2.4E-04)}\\
				\specialrule{0em}{1pt}{1pt}
				&IMOP5 & 7.955E-01(5.2E-04)$^\dagger$ & 8.147E-01(4.3E-03)$^\dagger$ & \textbf{8.280E-01(1.0E-03)}$^\dagger$ & 8.217E-01(3.5E-03)$^\dagger$ & 8.247E-01(3.0E-03)\\
				\specialrule{0em}{1pt}{1pt}
				&IMOP6 & 4.201E-01(4.6E-04)$^\dagger$ & 4.274E-01(9.9E-04)$^\dagger$ & 4.299E-01(7.8E-04)$^\dagger$ & 4.323E-01(6.0E-04)$^\dagger$  & \textbf{4.324E-01(5.5E-04)}\\
				\hline
				\specialrule{0em}{1pt}{1pt}
				\multicolumn{2}{c}{$ +/=/- $ }& \textbf{$32/2/2$} &  \textbf{$29/3/4$} & \textbf{$22/5/9$} & \textbf{$20/13/3$} &  \\
				\hline
			\end{tabular}{}
			\begin{footnotesize}				
				$'\dagger'$ indicates that ATM-MOEA/D is of statistically significant difference from the corresponding peer algorithm at a 0.05 level by Wilcoxon's rank sum test. The best mean for each case is highlighted in boldface. The symbols $+, = $, and $- $ indicate that the results of ATM-MOEA/D are significantly better than, worse than, and equivalent to the corresponding peer algorithm.
			\end{footnotesize}
		\end{threeparttable}
	\end{footnotesize}
\end{table*}

\subsection{Experimental results}

Table~\ref{IGD-ATM-MOEA/D} and Table~\ref{HV-ATM-MOEA/D} show the IGD and HV results regarding the mean and standard deviation (SD) values of ATM-MOEA/D and the other four peer algorithms. 
The better mean for each problem was highlighted in boldface. 
To have statistically sound conclusions, 
the Wilcoxon’s rank sum test~\cite{Derrac2011} at a 0.05 significance level was used to test the significance of the differences between the results obtained by ATM-MOEA/D and the four peer algorithms.


From the tables, we can see that ATM-MOEA/D ranks the best overall compared with the four peer algorithms, both on IGD and HV. For IGD, ATM-MOEA/D outperforms MOEA/D, DEA-GNG, RVEA-iGNG, and AdaW on 31, 35, 30, and 17 out of the 36 test problems, respectively. In contrast, MOEA/D, DEA-GNG, RVEA-iGNG, and AdaW win only on 2, 0, 5, and 3 test problems. For HV, ATM-MOEA/D outperforms MOEA/D, DEA-GNG, RVEA-iGNG, and AdaW on 32, 29, 22, and 20 test problems, respectively. In contrast, MOEA/D, DEA-GNG, RVEA-iGNG, and AdaW win only on 2, 4, 9, and 3 test problems.

\begin{figure*}[tbp]
	\begin{center}
		\footnotesize
		\hspace*{-20pt}\begin{tabular}{@{}c@{}c@{}c@{}c@{}c@{}}
			\includegraphics[scale=0.335]{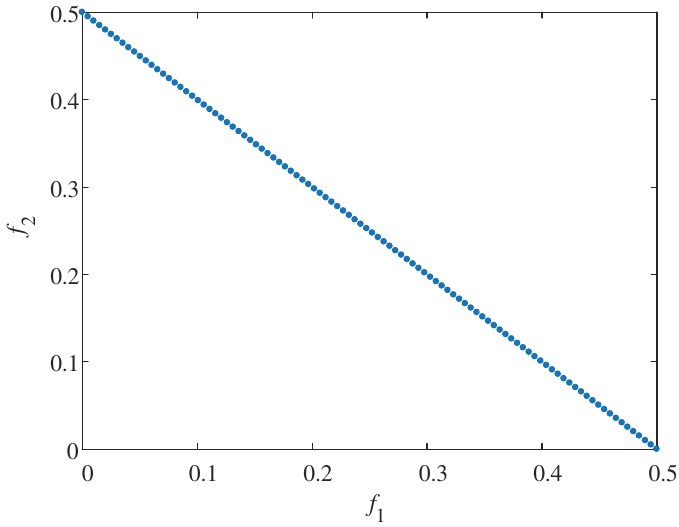}&
			\includegraphics[scale=0.335]{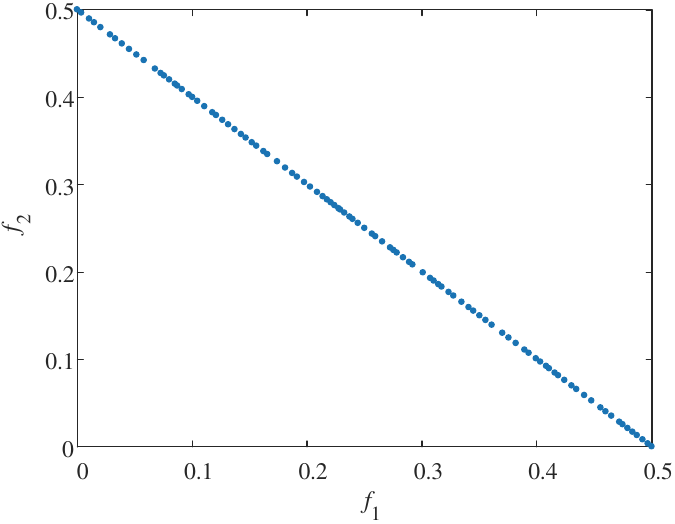}&
			\includegraphics[scale=0.335]{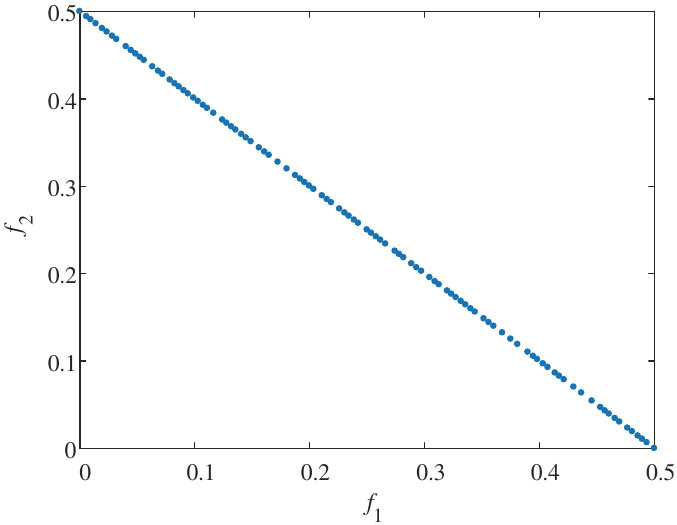}&
			\includegraphics[scale=0.335]{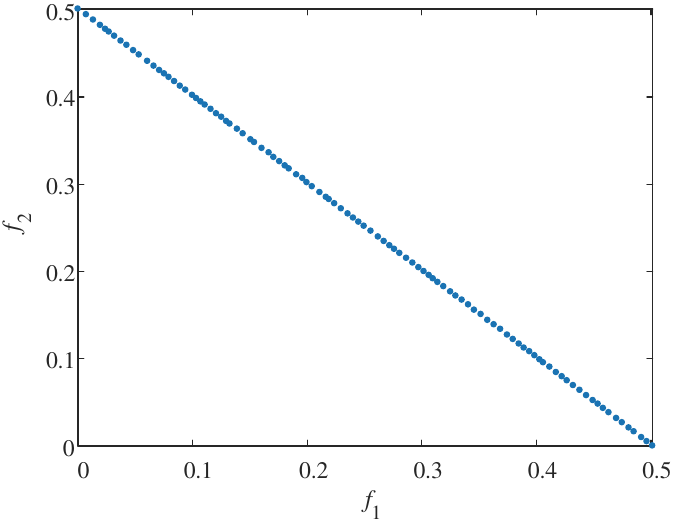}&
			\includegraphics[scale=0.335]{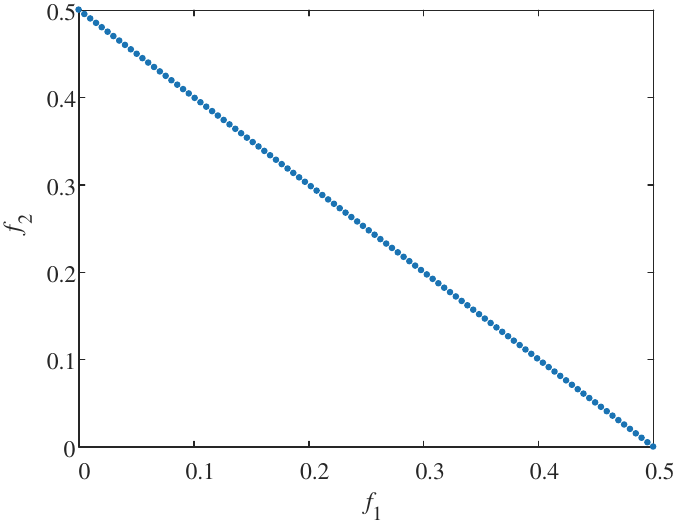}\\
			(a) MOEA/D & (b) DEA-GNG & (c) RVEA-iGNG & (d) AdaW  & (e) ATM-MOEA/D \\
			
		\end{tabular}
	\end{center}
	\caption{The final solution set of the five algorithms on the 2-objective DTLZ1.}
	\label{2objDTLZ1}
\end{figure*}

\begin{figure*}[tbp]
	\begin{center}
		\footnotesize
		\hspace*{-20pt}\begin{tabular}{@{}c@{}c@{}c@{}c@{}c@{}}
			\includegraphics[scale=0.335]{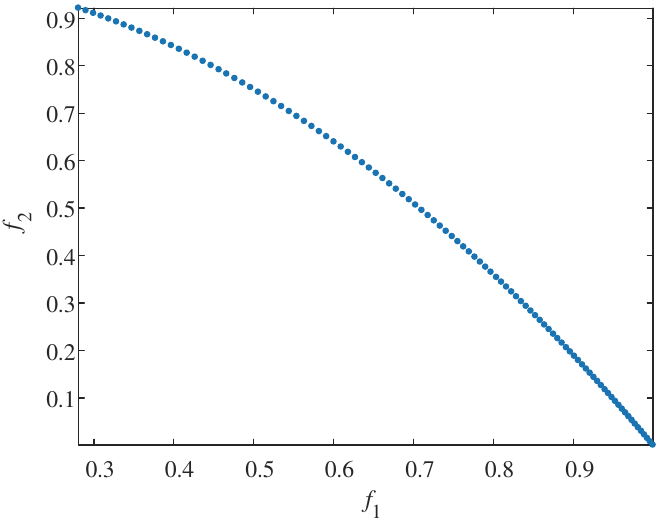}&
			\includegraphics[scale=0.335]{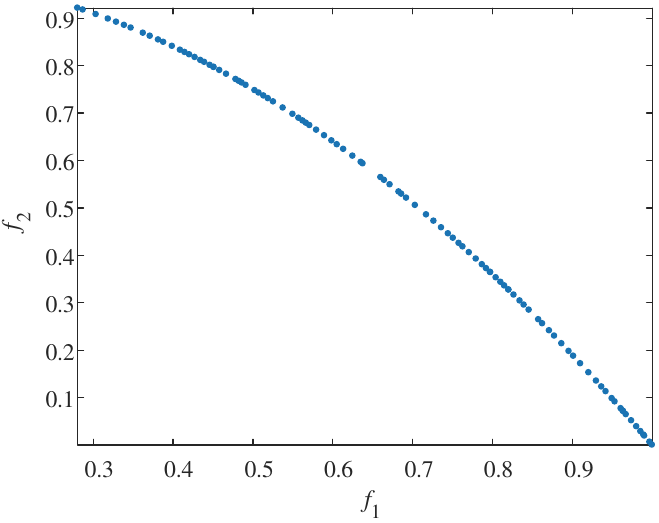}&
			\includegraphics[scale=0.335]{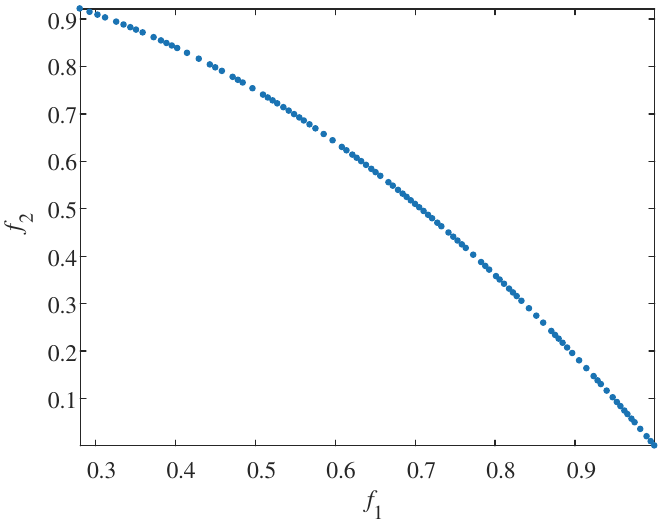}&
			\includegraphics[scale=0.335]{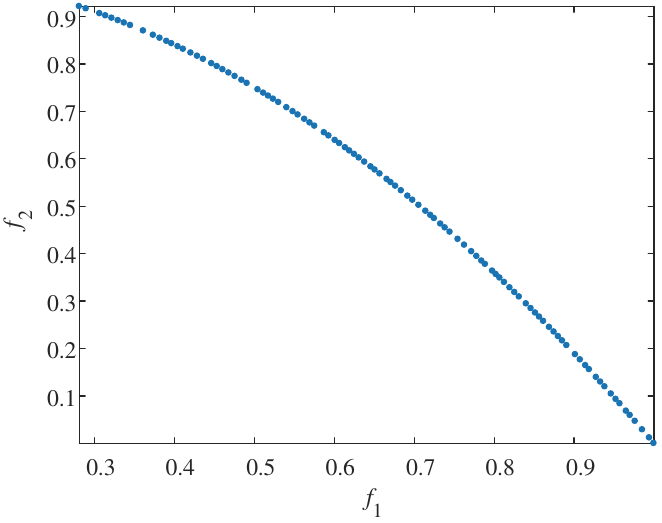}&
			\includegraphics[scale=0.335]{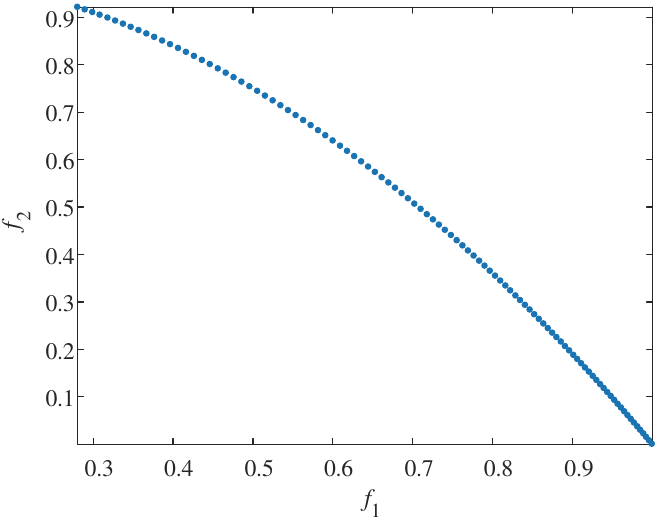}\\
			(a) MOEA/D & (b) DEA-GNG & (c) RVEA-iGNG & (d) AdaW  & (e) ATM-MOEA/D \\
		\end{tabular}
	\end{center}
	\caption{The final solution set of the five algorithms on ZDT6.}
	\label{ZDT6}
\end{figure*}

\subsubsection{On problems with a regular Pareto front}
\begin{figure*}[tbp]
	\begin{center}
		\footnotesize
		\hspace*{-20pt}\begin{tabular}{@{}c@{}c@{}c@{}c@{}c@{}}
			\includegraphics[scale=0.32]{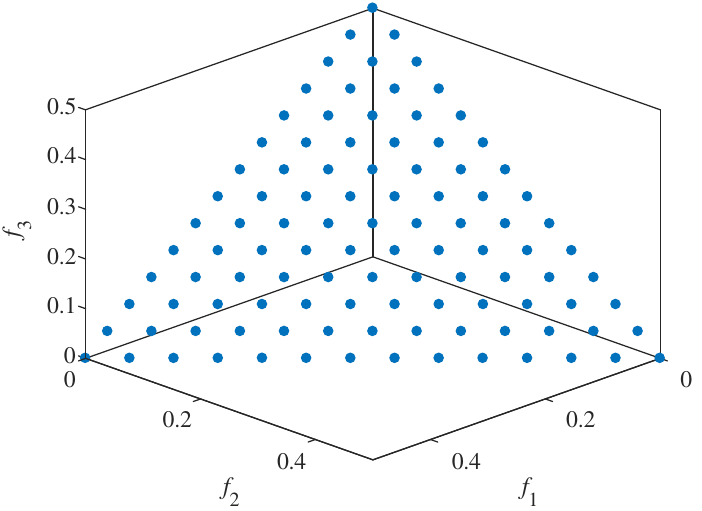}&
			\includegraphics[scale=0.32]{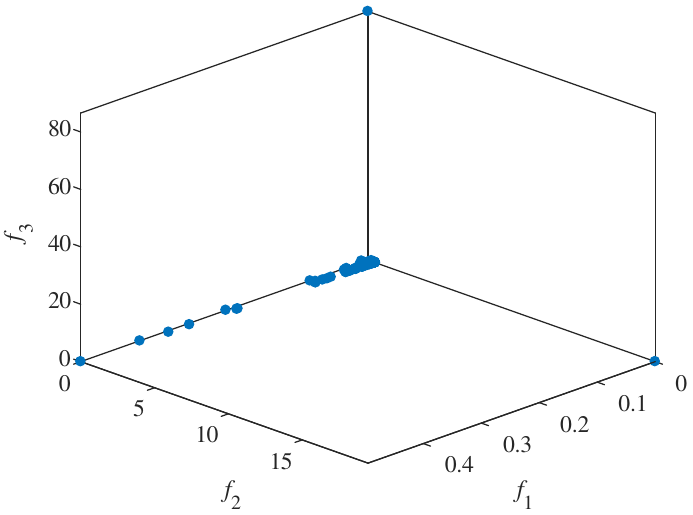}&
			\includegraphics[scale=0.32]{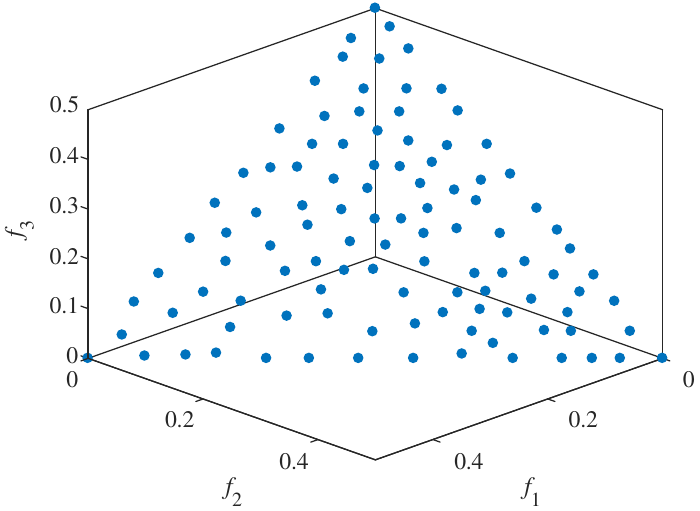}&
			\includegraphics[scale=0.32]{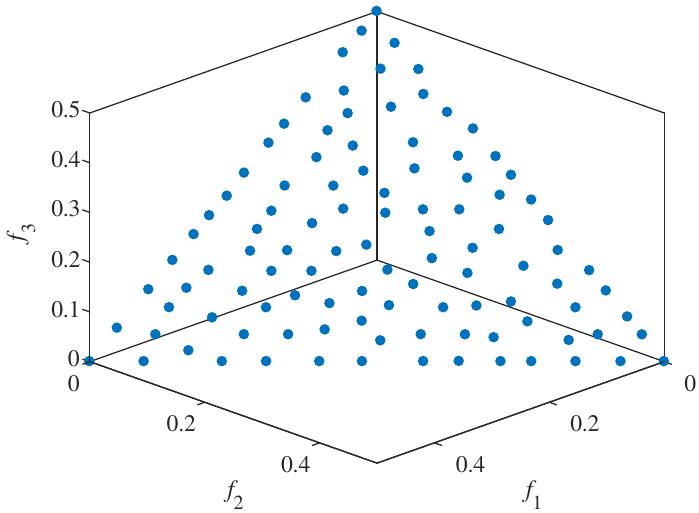}&
			\includegraphics[scale=0.32]{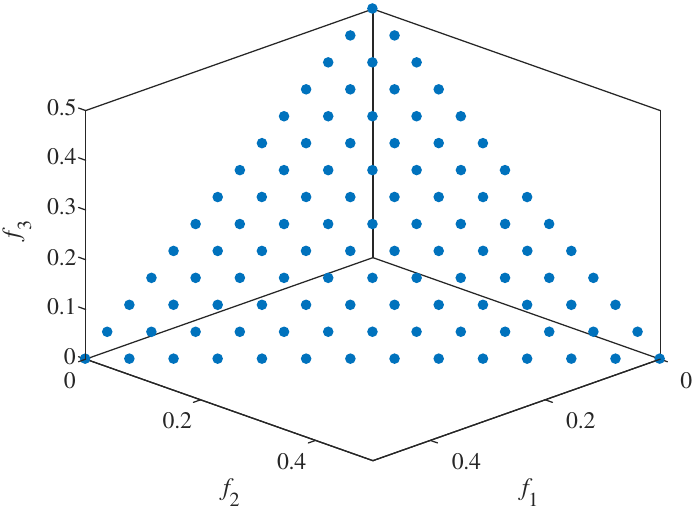}\\
(a) MOEA/D & (b) DEA-GNG & (c) RVEA-iGNG & (d) AdaW  & (e) ATM-MOEA/D \\
			
		\end{tabular}
	\end{center}
	\caption{The final solution set of the five algorithms on the 3-objective DTLZ1.}
	\label{3objDTLZ1}
\end{figure*}

\begin{figure*}[tbp]
	\begin{center}
		\footnotesize
		\hspace*{-20pt}\begin{tabular}{@{}c@{}c@{}c@{}c@{}c@{}}
			\includegraphics[scale=0.33]{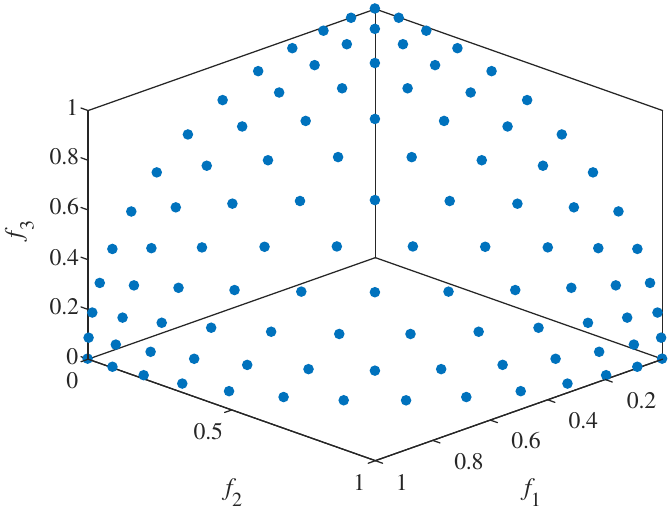}&
			\includegraphics[scale=0.33]{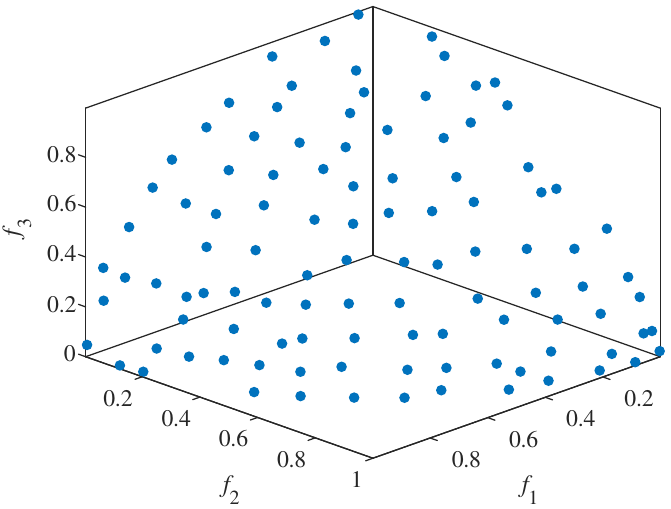}&
			\includegraphics[scale=0.33]{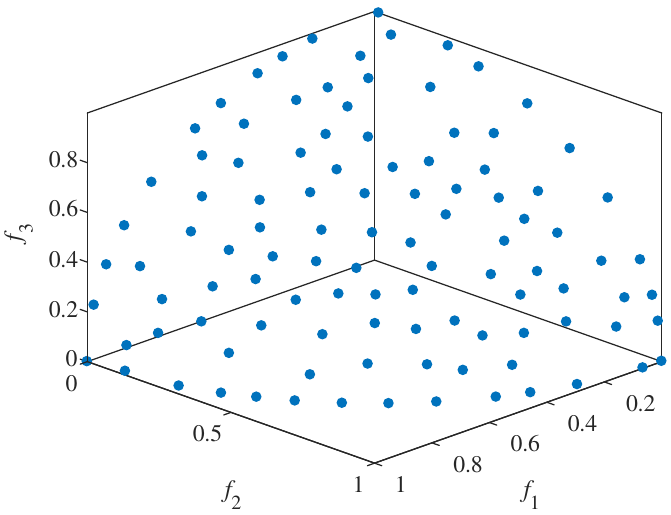}&
			\includegraphics[scale=0.33]{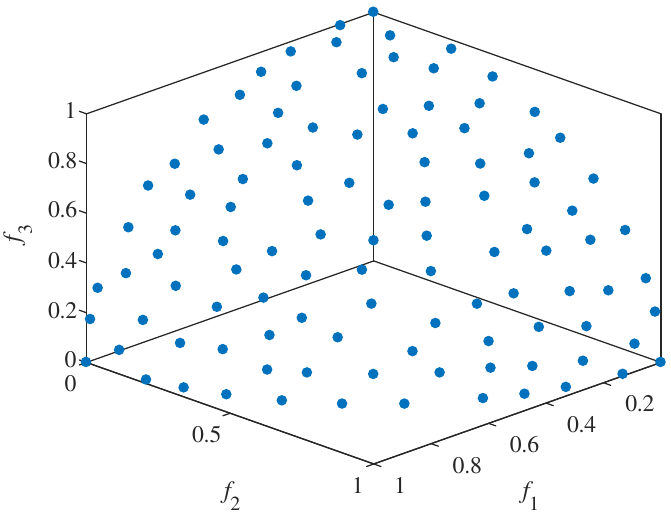}&
			\includegraphics[scale=0.33]{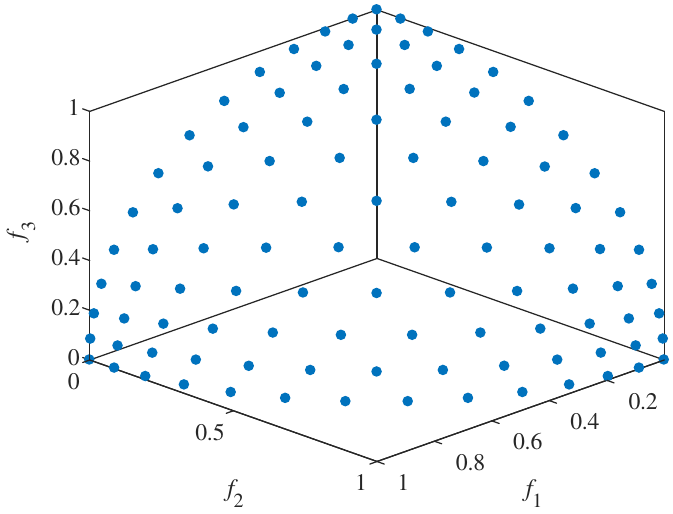}\\
            (a) MOEA/D & (b) DEA-GNG & (c) RVEA-iGNG & (d) AdaW  & (e) ATM-MOEA/D \\	
		\end{tabular}
	\end{center}
	\caption{The final solution set of the five algorithms on the 3-objective DTLZ2.}
	\label{3objDTLZ2}
\end{figure*}

\begin{figure*}[tbp]
	\begin{center}
		\footnotesize
		\hspace*{-20pt}\begin{tabular}{@{}c@{}c@{}c@{}c@{}c@{}}
			\includegraphics[scale=0.335]{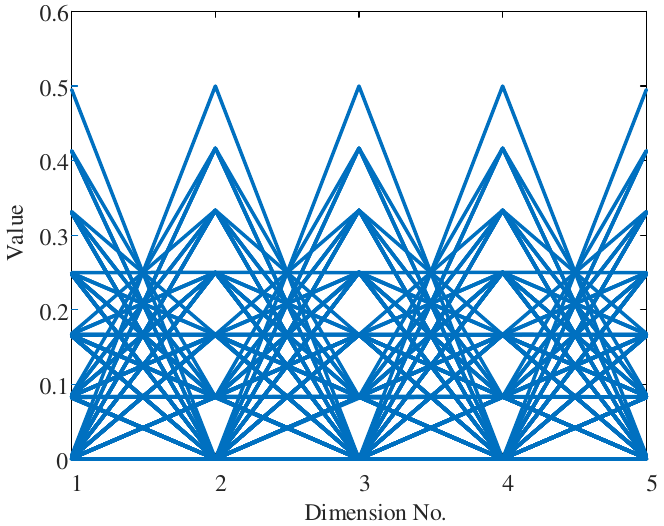}&
			\includegraphics[scale=0.335]{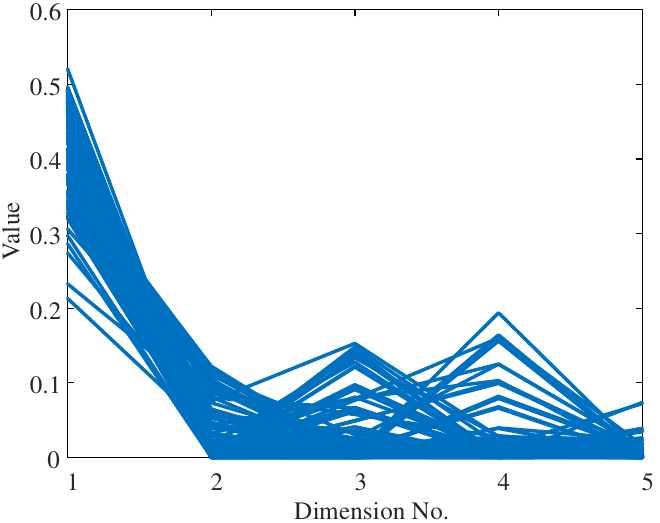}&
			\includegraphics[scale=0.335]{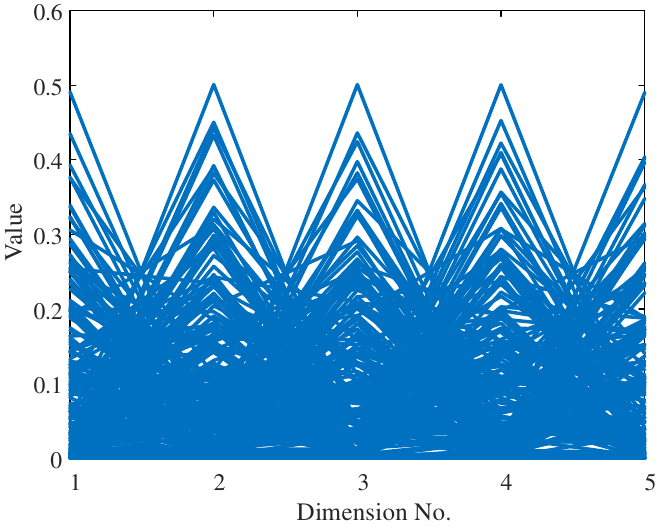}&
			\includegraphics[scale=0.335]{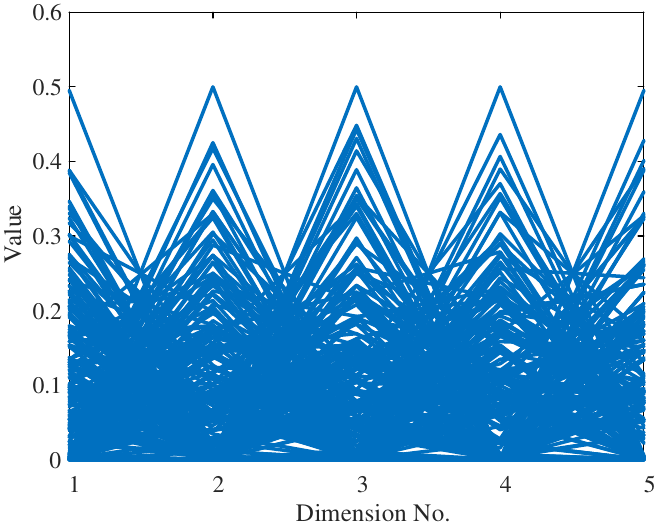}&
			\includegraphics[scale=0.335]{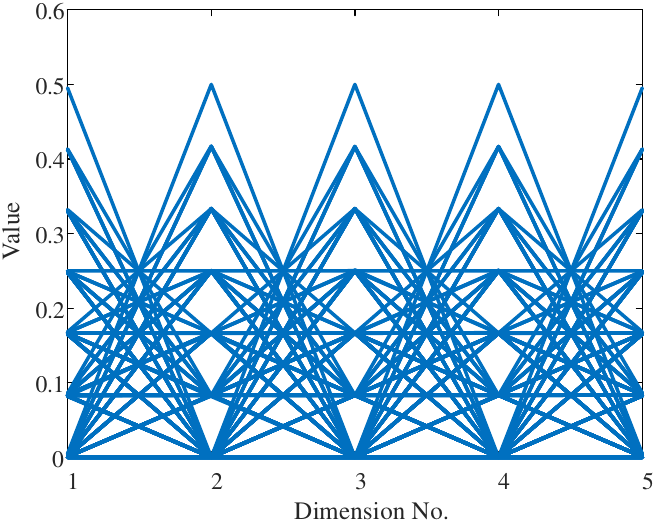}\\
			(a) MOEA/D & (b) DEA-GNG & (c) RVEA-iGNG & (d) AdaW  & (e) ATM-MOEA/D \\	
		\end{tabular}
	\end{center}
	\caption{The final solution set of the five algorithms on the 5-objective DTLZ2.}
	\label{5objDTLZ1}
\end{figure*}

ATM-MOEA/D performs well on problems with regular Pareto front. 
We selected 10 test problems with a regular Pareto front and evaluated the performance of the five algorithms using two indicators: IGD and HV.
In comparison with the three weight adaptation algorithms (DEA-GNG, RVEA-iGNG, and AdaW), 
ATM-MOEA/D exhibits better performance on problems with a regular Pareto front.
According to Tables~\ref{IGD-ATM-MOEA/D} and~\ref{HV-ATM-MOEA/D}, 
for IGD, ATM-MOEA/D outperforms DEA-GNG, RVEA-iGNG, and AdaW on 10,9,10 out of the 10 test problems, respectively. 
For HV, ATM-MOEA/D outperforms DEA-GNG, RVEA-iGNG, and AdaW on 8,7,9 out of the 10 test problems, respectively.
Figures~\ref{2objDTLZ1}--\ref{5objDTLZ1} illustrate the final solution sets obtained by the five algorithms on the representative problems with regular Pareto fronts. As shown in the figures, ATM-MOEA/D outperforms the other three algorithms on both the 2-objective problems (DTLZ1, ZDT6) and the 3-objective problems (DTLZ1, DTLZ2) with its solutions being well distributed on a line or a face. 
The solutions of the other three weight adaptation algorithms are not so uniform as those of ATM-MOEA/D.
For the 5-objective DTLZ2, Figure~\ref{5objDTLZ1} plots the final solution sets using parallel coordinates.
We can not conclude the distribution difference of the algorithms by the parallel coordinates plots. 
All five algorithms appear to work well, although there exist several solutions of AdaW, RVEA-iGNG, and DEA-GNG that do not fully converge to the Pareto front.
However, according to the IGD and HV results in Tables~\ref{IGD-ATM-MOEA/D} and~\ref{HV-ATM-MOEA/D}, ATM-MOEA/D performs better than the other three weight adaptation algorithms on the 5-objective DTLZ1 and DTLZ2.
The reasons why ATM-MOEA/D outperforms the three weight adjustment algorithms on the problems with a regular Pareto front are as follows. The three weight adaptation algorithms adapt the weights during the evolutionary process on problems with a regular Pareto front, which perturbs the uniform distribution of the initial weights of the algorithm, leading to an inconsistency between the weight distribution and the shape of the Pareto front.
Considering MOEA/D, which uses fixed weights during the evolutionary process, 
both ATM-MOEA/D and MOEA/D perform similarly in terms of the diversity of solutions on the problems with a regular Pareto front.
As shown in Figures~\ref{2objDTLZ1}--\ref{5objDTLZ1}, 
both ATM-MOEA/D and MOEA/D are effective on these problems.
This is because fixed weights help to obtain a more uniform solution set for regular Pareto fronts.

\subsubsection{On problems with an irregular Pareto front}
We selected 26 test problems with an irregular Pareto front and evaluated the performance of the five algorithms using two indicators: IGD and HV. The results are presented in Tables~\ref{IGD-ATM-MOEA/D} and~\ref{HV-ATM-MOEA/D}. 
In terms of IGD, ATM-MOEA/D outperforms MOEA/D, DEA-GNG, RVEA-iGNG, and AdaW on 26, 25, 21, and 7 out of the 26 test problems, respectively. In contrast, MOEA/D, DEA-GNG, RVEA-iGNG, and AdaW win only on 0, 0, 5, and 3 test problems. Similarly, for HV, ATM-MOEA/D outperforms MOEA/D, DEA-GNG, RVEA-iGNG, and AdaW on 26, 21, 15, and 11 out of the 26 test problems, respectively. In contrast, MOEA/D, DEA-GNG, RVEA-iGNG, and AdaW win only on 0, 3, 8, and 3 test problems.

\begin{figure*}[tbp]
	\begin{center}
		\footnotesize
		\hspace*{-20pt}\begin{tabular}{@{}c@{}c@{}c@{}c@{}c@{}}
			\includegraphics[scale=0.325]{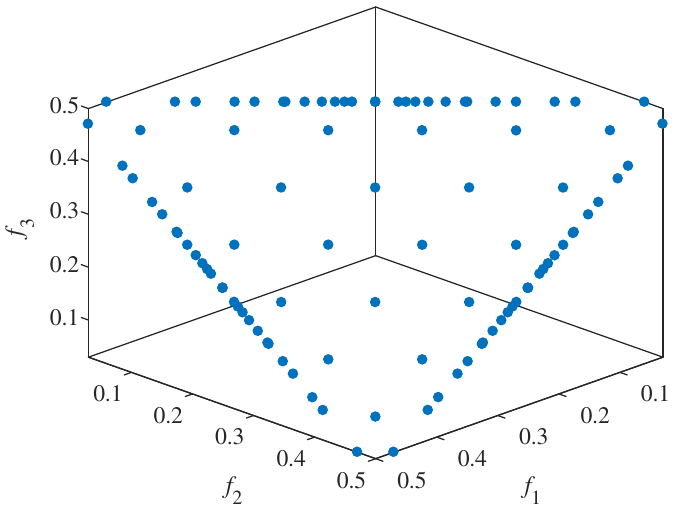}&
			\includegraphics[scale=0.325]{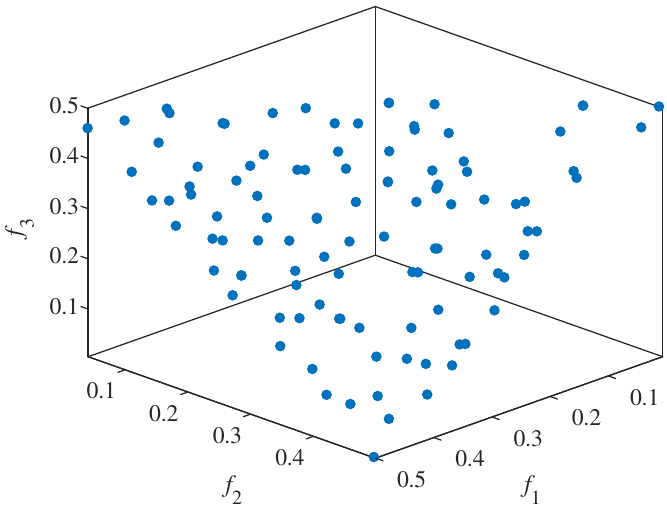}&
			\includegraphics[scale=0.325]{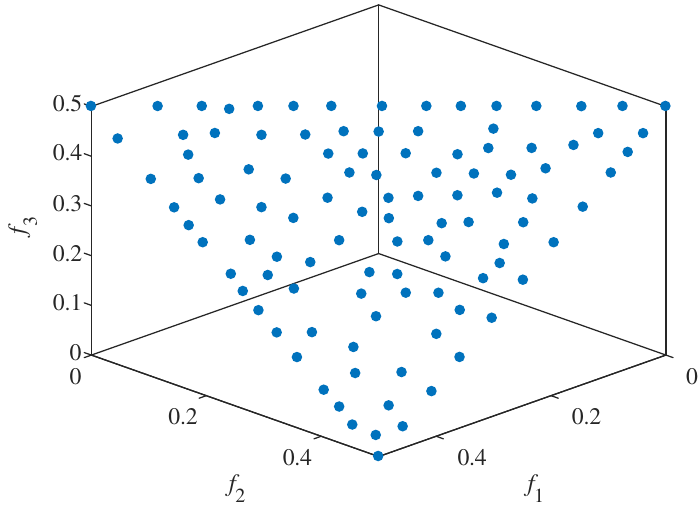}&
			\includegraphics[scale=0.325]{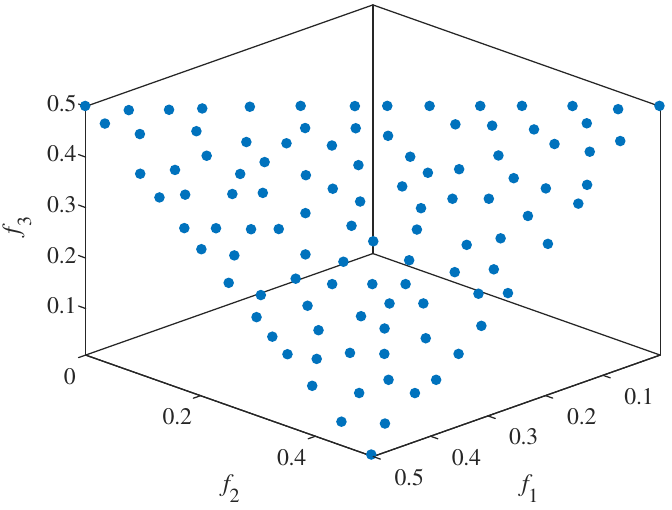}&
			\includegraphics[scale=0.325]{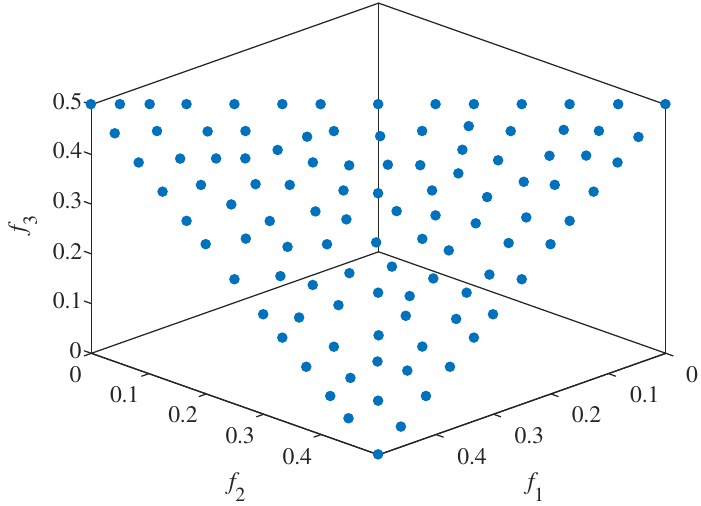}\\
			(a) MOEA/D & (b) DEA-GNG & (c) RVEA-iGNG & (d) AdaW  & (e) ATM-MOEA/D \\
		\end{tabular}
	\end{center}
	\caption{The final solution set of the five algorithms on the 3-objective IDTLZ1.}
	\label{3objIDTLZ1}
\end{figure*}


\begin{figure*}[tbp]
	\begin{center}
		\footnotesize
		\hspace*{-20pt}\begin{tabular}{@{}c@{}c@{}c@{}c@{}c@{}}
			\includegraphics[scale=0.32]{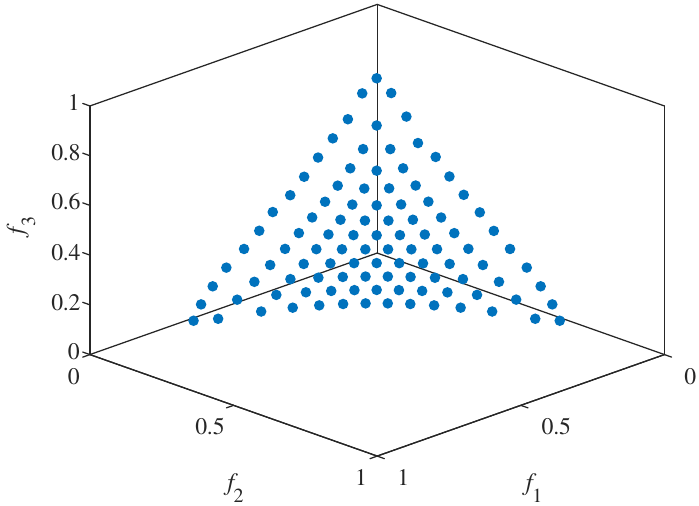}&
			\includegraphics[scale=0.32]{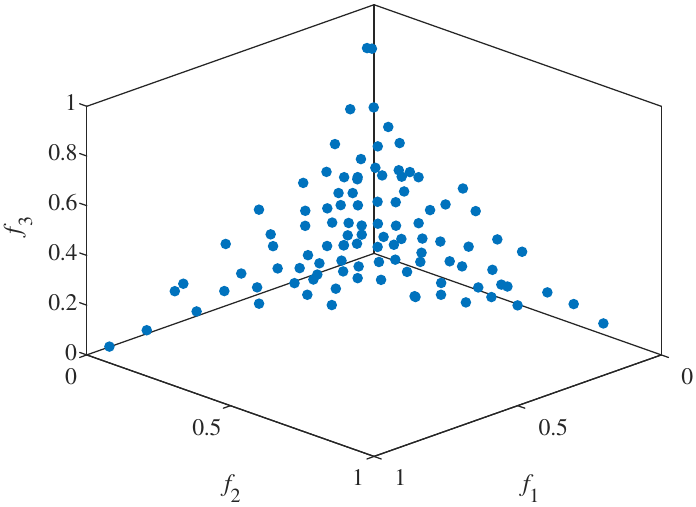}&
			\includegraphics[scale=0.32]{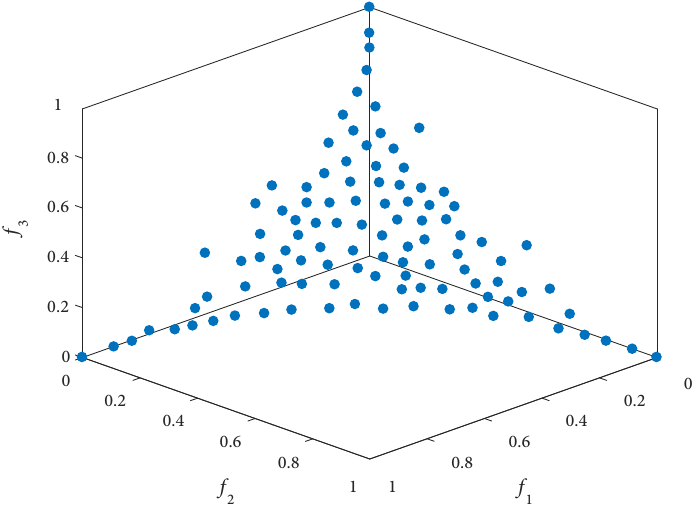}&
			\includegraphics[scale=0.32]{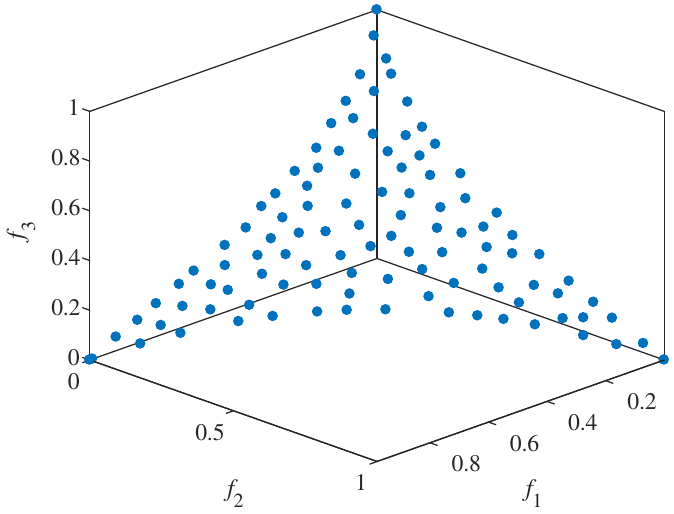}&
			\includegraphics[scale=0.32]{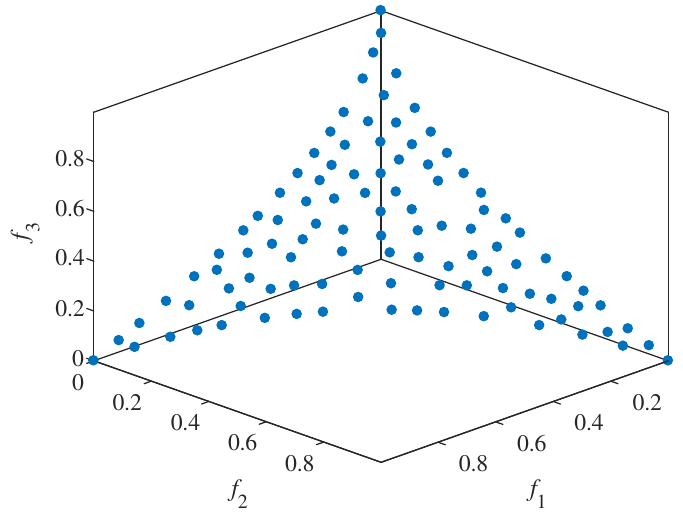}\\
			(a) MOEA/D & (b) DEA-GNG & (c) RVEA-iGNG & (d) AdaW  & (e) ATM-MOEA/D \\
		\end{tabular}
	\end{center}
	\caption{The final solution set of the five algorithms on the 3-objective CDTLZ2.}
	\label{3objCDTLZ2}
\end{figure*}

\begin{figure*}[tbp]
	\begin{center}
		\footnotesize
		\hspace*{-20pt}\begin{tabular}{@{}c@{}c@{}c@{}c@{}c@{}}
			\includegraphics[scale=0.325]{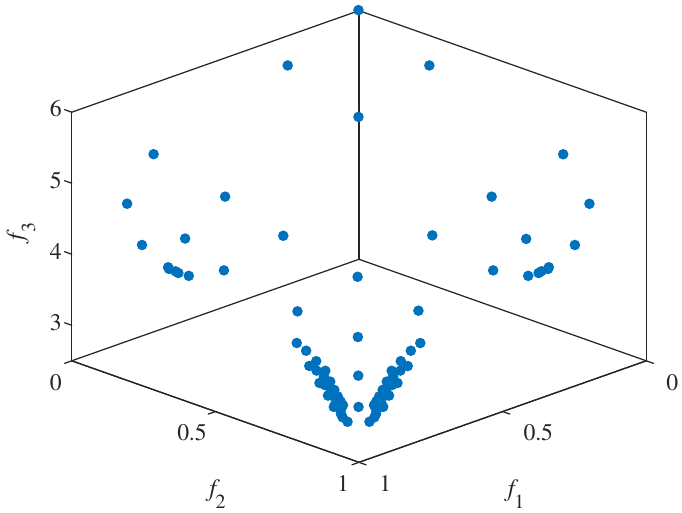}&
			\includegraphics[scale=0.325]{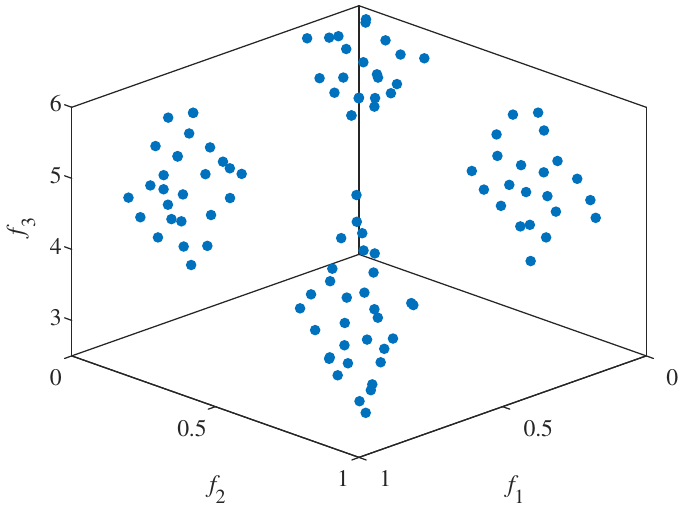}&
			\includegraphics[scale=0.325]{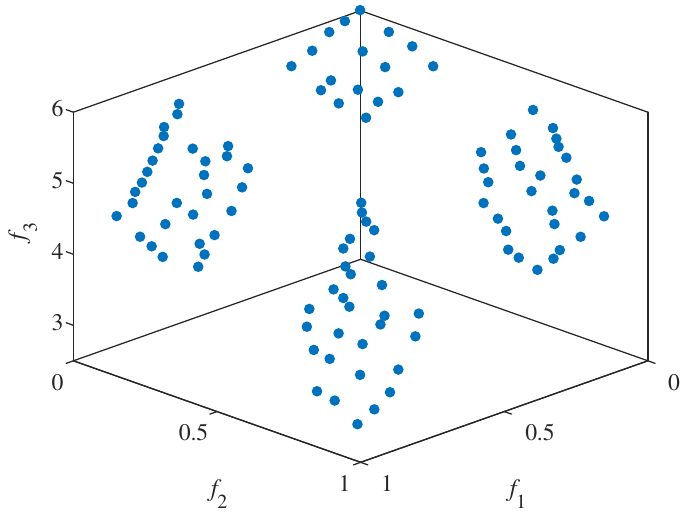}&
			\includegraphics[scale=0.325]{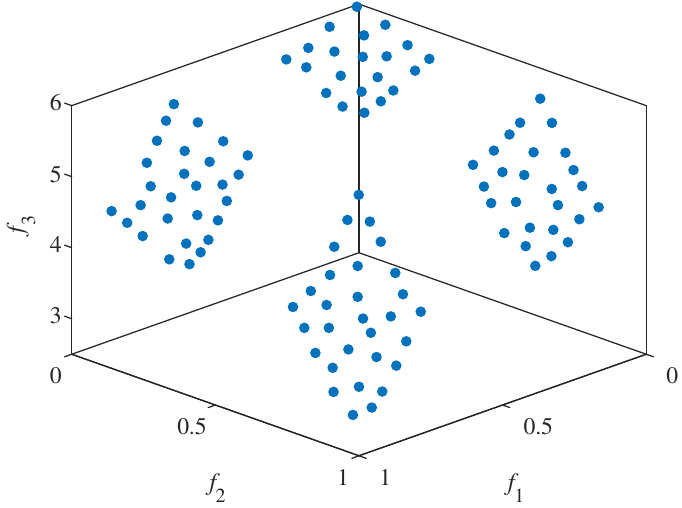}&
			\includegraphics[scale=0.325]{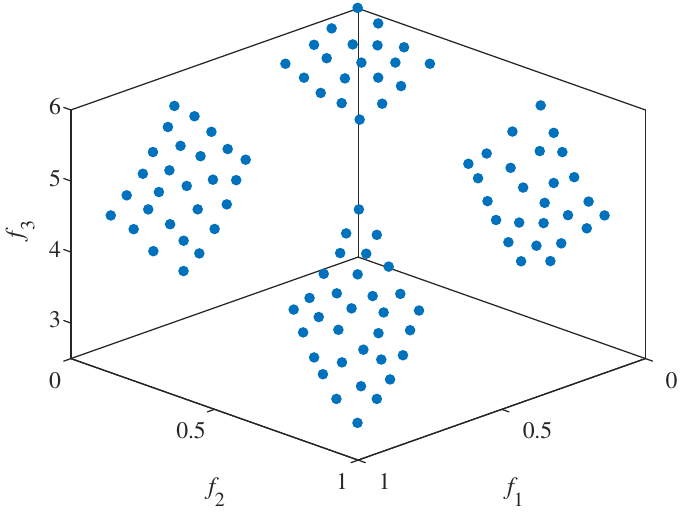}\\
			(a) MOEA/D & (b) DEA-GNG & (c) RVEA-iGNG & (d) AdaW  & (e) ATM-MOEA/D \\
		\end{tabular}
	\end{center}
	\caption{The final solution set of the five algorithms on the 3-objective DTLZ7.}
	\label{3objDTLZ7}
\end{figure*}

\begin{figure*}[tbp]
	\begin{center}
		\footnotesize
		\hspace*{-20pt}\begin{tabular}{@{}c@{}c@{}c@{}c@{}c@{}}
			\includegraphics[scale=0.31]{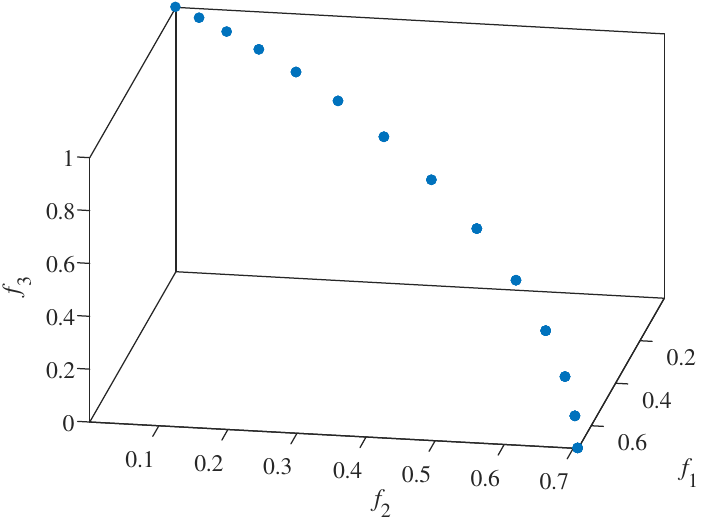}&
			\includegraphics[scale=0.31]{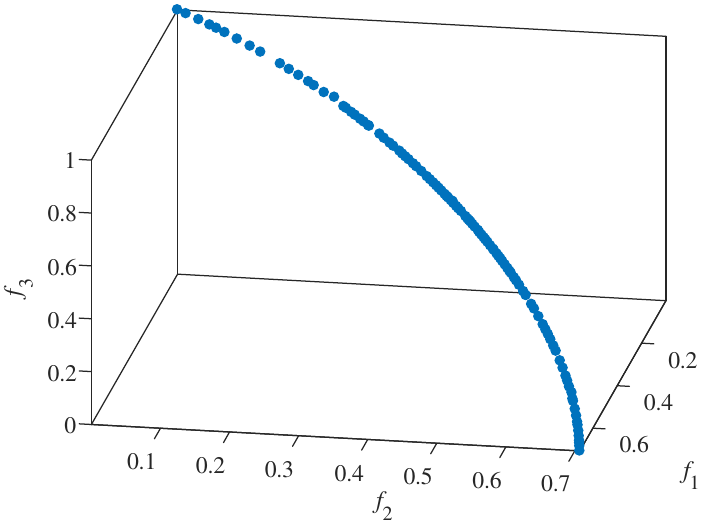}&
			\includegraphics[scale=0.31]{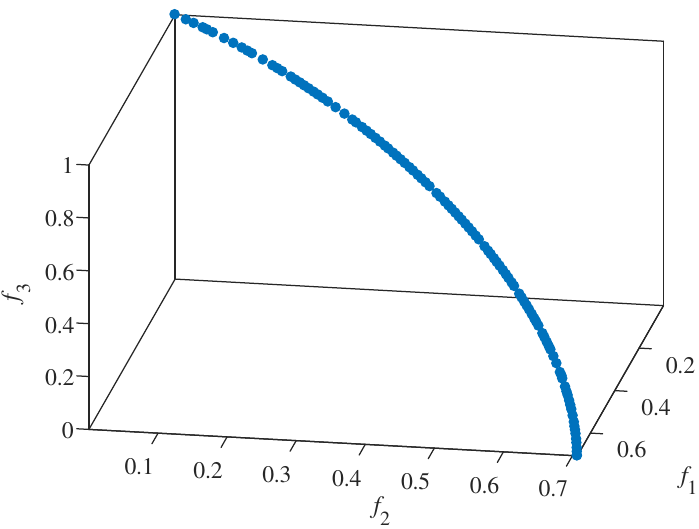}&
			\includegraphics[scale=0.31]{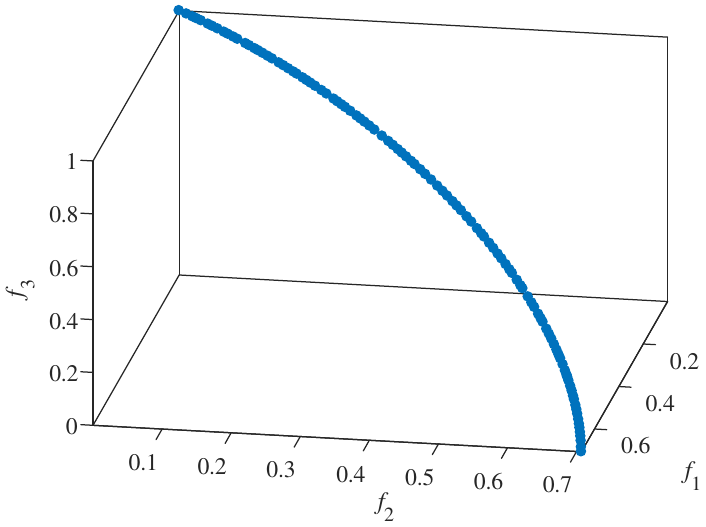}&
			\includegraphics[scale=0.31]{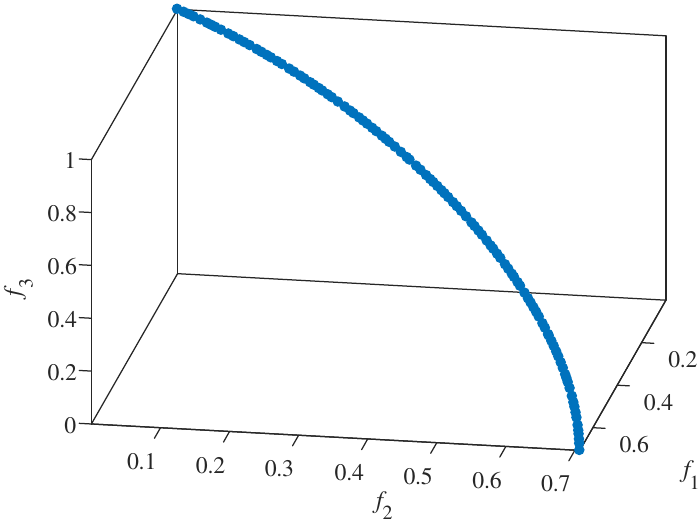}\\
			(a) MOEA/D & (b) DEA-GNG & (c) RVEA-iGNG & (d) AdaW  & (e) ATM-MOEA/D \\
		\end{tabular}
	\end{center}
	\caption{The final solution set of the five algorithms on the 3-objective DTLZ5.}
	\label{3objDTLZ5}
\end{figure*}


\begin{figure*}[tbp]
	\begin{center}
		\footnotesize
		\hspace*{-20pt}\begin{tabular}{@{}c@{}c@{}c@{}c@{}c@{}}
			\includegraphics[scale=0.315]{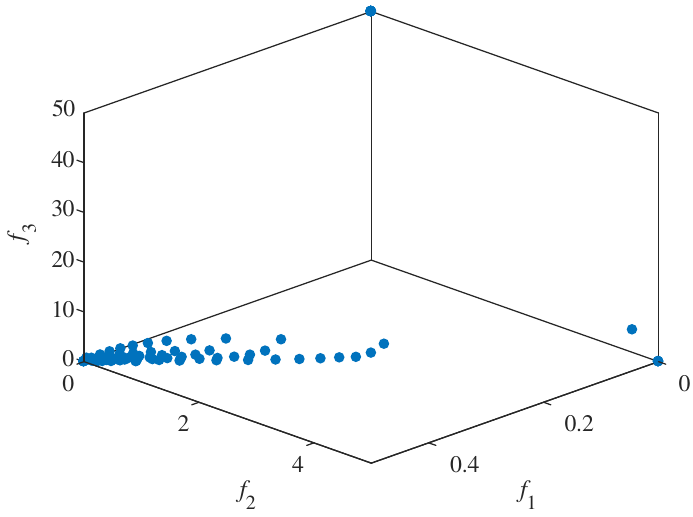}&
			\includegraphics[scale=0.315]{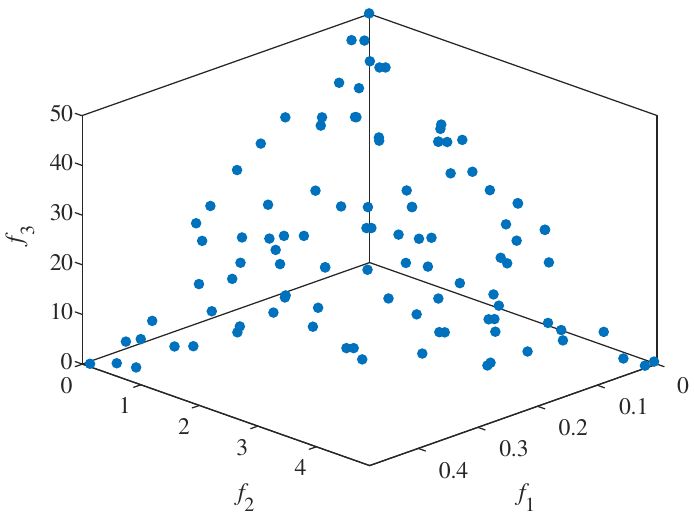}&
			\includegraphics[scale=0.315]{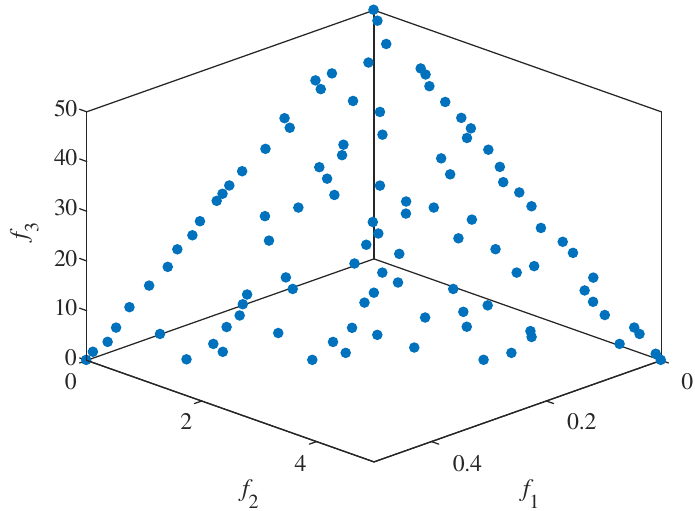}&
			\includegraphics[scale=0.315]{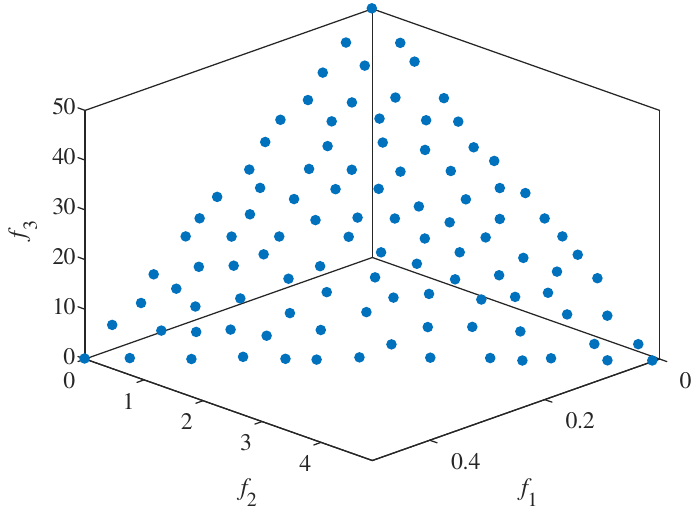}&
			\includegraphics[scale=0.315]{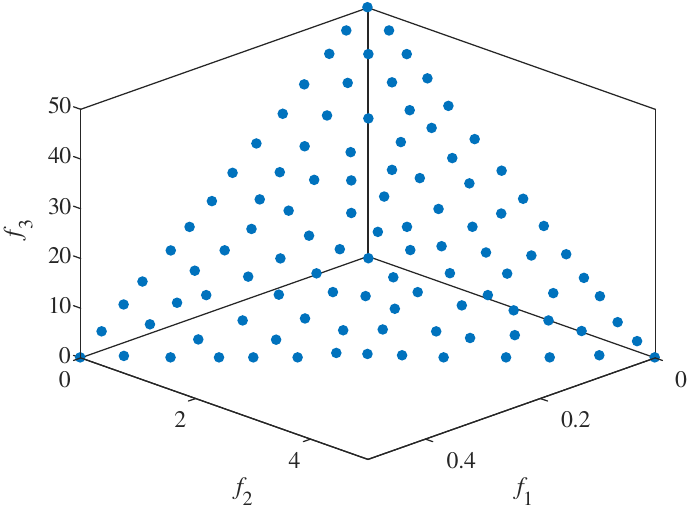}\\
			(a) MOEA/D & (b) DEA-GNG & (c) RVEA-iGNG & (d) AdaW  & (e) ATM-MOEA/D \\
		\end{tabular}
	\end{center}
	\caption{The final solution set of the five algorithms on the 3-objective SDTLZ1.}
	\label{3objSDTLZ1}
\end{figure*}


To gain a visual understanding of the search behaviour of the five algorithms, 
we plot one representative problem from each of the five problem categories with irregular Pareto fronts: 
the 3-objective IDTLZ1 for inverted Pareto fronts, CDTLZ2 for high-nonlinear Pareto fronts, the 3-objective DTLZ7 for disconnect Pareto fronts, the 3-objective DTLZ5 for degenerate Pareto fronts, and the 3-objective SDTLZ1 for scaled Pareto fronts.
For the 3-objective IDTLZ1 (refer to Figure~\ref{3objIDTLZ1}), solutions obtained by MOEA/D are mainly located on the boundary part of the Pareto front. 
Solutions obtained by DEA-GNG and RVEA-iGNG, compared with ATM-MOEA/D, are not uniformly distributed on the Pareto front. 
DEA-GNG has multiple solutions clustered together, while RVEA-iGNG has blank regions in its solution set. AdaW's solutions are not as uniform as that of ATM-MOEA/D.
For the 3-objective CDTLZ2 (refer to Figure~\ref{3objCDTLZ2}), MOEA/D generates very few solutions in the end. DEA-GNG has difficulty in preserving boundary solutions and exhibits clustering behaviour. The uniformity of the solution sets generated by RVEA-iGNG and AdaW is inferior to that of ATM-MOEA/D.
For the 3-objective DTLZ7 (refer to Figure~\ref{3objDTLZ7}), MOEA/D generates solutions that are clustered on the boundary. DEA-GNG has a similar issue with preserving boundary solutions and also exhibits clustering behaviour. The uniformity of the solution sets generated by RVEA-iGNG and AdaW is inferior to that of ATM-MOEA/D.
For the 3-objective DTLZ5 (refer to Figure~\ref{3objDTLZ5}), MOEA/D obtains a small number of solutions in the end, due to the existence of multiple weights corresponding to a single solution. 
The uniformity of the solutions obtained by DEA-GNG and RVEA-iGNG algorithms is inferior to that of ATM-MOEA/D; 
they have gaps in the upper part of the Pareto front. 
For the 3-objective SDTLZ1 (refer to Figure~\ref{3objSDTLZ1}), MOEA/D tends to generate solutions that are concentrated in one corner of the Pareto front due to the lack of normalisation. Although DEA-GNG and RVEA-iGNG cover the whole Pareto front, solutions obtained by DEA-GNG have poor uniformity, and most solutions obtained by RVEA-iGNG are on the boundary. ATM-MOEA/D has better uniformity than AdaW in terms of solution distribution. In summary, ATM-MOEA/D outperforms the other algorithms in terms of IGD and HV on most of the problems with irregular Pareto fronts. 
It generates more uniformly distributed solution sets for these problems.

\section{CONCLUSION}
For a decomposition-based MOEA, 
the ability to identify the nature of the given problem's Pareto front and adjust the weights accordingly during the evolution is helpful in finding a well-distributed Pareto front approximation. 
Along this line, 
this paper proposed a weight adaptation trigger mechanism to dynamically adapt the weights based on the status of the evolutionary search. 
Experimental results demonstrated that ATM-MOEA/D is able to maintain well-distributed solutions on both problems with regular Pareto fronts and problems with irregular Pareto fronts.
In addition, an extra benefit of ATM-MOEA/D is that it can reduce the frequency of the weight change, thus enhancing the convergence of the search. 

Despite the promising results in general, there are some issues of the proposed algorithm worth noting. 
First, ATM-MOEA/D requires settings of several parameters such as the number of generations for determining the evolutionary stagnation, the niche size for consistency detection between the archive and population, and the size of the archive. 
While we kept these parameters constant on all the test problems, 
individual settings for specific problems may improve the results. 
Another issue is that the algorithm may become less effective when the archive based on the Pareto dominance and crowdedness criteria fails to accurately reflect the problem's Pareto front, such as on MOPs with many ``dominance resistance solutions''. 
In such problems, when the evolution stagnates, the archive may not well reflect the shape of the problem's Pareto front. 
Consequently, the accuracy of the estimation of the problem's shape and 
the timing of adapting the weights will largely be affected. 
This issue is relevant to all algorithms that utilise an archive to assist in weight adaptation, and we believe it to be an important topic and aim to address it in the near future.



\bibliographystyle{IEEEtran}
\bibliography{IEEEabrv,bibfile}

\end{document}